\RequirePackage{fix-cm}
\documentclass[twocolumn]{svjour3}          
\usepackage{graphicx}
\usepackage{amsfonts}
\usepackage{amsmath}
\usepackage{mathtools}
\usepackage{bbm}
\usepackage[inline]{enumitem}
\usepackage{tabularx}
\usepackage{multirow}
\usepackage{url}
\usepackage{booktabs}
\usepackage[caption=false,font=footnotesize,labelfont=rm,textfont=rm]{subfig}
\usepackage{diagbox}
\usepackage{microtype}
\usepackage[numbers,sort]{natbib}

\usepackage{titlesec}
\titleformat{\paragraph}
{\normalfont\normalsize\bfseries} 
{\theparagraph} 
{1em} 
{} 
\titlespacing*{\paragraph}
{0pt} 
{1ex plus .1ex minus .2ex} 
{0.5ex plus .1ex} 

\usepackage{siunitx}
\usepackage{threeparttable}

\newcommand{\modelname}{\textsl{CroTad}}

\providecommand\given{} 
\newcommand\SetSymbol[1][]{
   \nonscript\,#1\vert \allowbreak \nonscript\,\mathopen{}}
\DeclarePairedDelimiterX\Set[1]{\lbrace}{\rbrace}
 { \renewcommand\given{\SetSymbol[\delimsize]} #1 }
\begin{document}

\sloppy

\title{CroTad: A Contrastive Reinforcement Learning Framework for Online Trajectory Anomaly Detection\thanks{Work done during employment by The Hong Kong Polytechnic University.}
}


\author{Rui Xue \and
        Dan He \and
        Fengmei Jin \and
        Chen Zhang \and
        Xiaofang Zhou
}


\institute{Rui Xue \at
              Independent Researcher \\
              \email{waldron.xue@outlook.com}           
           \and
           Dan He \at
           The University of Queensland \\
           \email{d.he@uq.edu.au}
           \and
           Fengmei Jin \at
           The Hong Kong Polytechnic University \\
           \email{fengmei.jin@outlook.com}
           \and
           Chen Zhang \\
           Corresponding Author \at
           The Hong Kong Polytechnic University \\
           \email{jason-c.zhang@polyu.edu.hk}
           \and
           Xiaofang Zhou \at
           The Hong Kong University of Science and Technology \\
           \email{zxf@cse.ust.hk}
    }

\date{Received: date / Accepted: date}

\maketitle

\begin{abstract}
Detecting trajectory anomalies is a vital task in modern Intelligent Transportation Systems (ITS), enabling the identification of unsafe, inefficient, or irregular travel behaviours. While deep learning has emerged as the dominant approach, several key challenges remain unresolved. First, sub-trajectory anomaly detection—capable of pinpointing the precise segments where anomalies occur—remains underexplored compared to whole-trajectory analysis. Second, many existing methods depend on carefully tuned thresholds, limiting their adaptability in real-world applications. Moreover, the irregular sampling of trajectory data and the presence of noise in training sets further degrade model performance, making it difficult to learn reliable representations of normal routes. To address these challenges, we propose a contrastive reinforcement learning framework for online trajectory anomaly detection, \modelname{}. Our method is threshold-free and robust to noisy, irregularly sampled data. By incorporating contrastive learning, \modelname{} learns to extract diverse normal travel patterns for different itineraries and effectively distinguish anomalous behaviours at both sub-trajectory and point levels. The detection module leverages deep reinforcement learning to perform online, real-time anomaly scoring, enabling timely and fine-grained identification of abnormal segments. Extensive experiments on two real-world datasets demonstrate the effectiveness and robustness of our framework across various evaluation scenarios.
\end{abstract}

\section{Introduction}

The proliferation of location-aware mobile devices, such as smartphones, has led to an explosion in spatial trajectory data generated by pedestrians, taxis, and public transit. This data forms the backbone of modern Intelligent Transportation Systems (ITS), enabling the discovery of mobility patterns that are critical for public safety \cite{deepcrime}, intelligent vehicle systems \cite{9083672}, traffic management, and urban planning \cite{wang2024traffic}. Moreover, trajectory analysis also supports traffic dynamics modeling \cite{hanDeepTEAEffectiveEfficient2022}, transport network optimization, and spatial-temporal data management \cite{chen2024deep}.


Detecting anomalies in trajectories is a critical task in this field, with widespread applications in real-world scenarios. Irregular trajectory patterns often indicate unsafe behavior, fraudulent activities, or violations of regulations. For example, anomaly detection techniques have been employed to identify cases of taxi fare manipulation and unauthorized route changes \cite{zhangIBATDetectingAnomalous2011, chenIBOATIsolationBasedOnline2013}. In Hong Kong, light buses are notorious for engaging in risky driving practices, underscoring the importance of real-time detection systems that enhance safety and bolster public trust. As Intelligent Transportation Systems (ITS) continue to evolve, the demand for accurate and scalable methods for trajectory anomaly detection becomes increasingly critical.

 Online trajectory anomaly detection involves identifying irregular movement patterns in real-time, where anomalies deviate from historical norms or predefined rules. Unlike offline methods that process data in batches, online systems continuously monitor trajectory streams and respond with minimal delay to ensure timely incident detection and response. Despite advances in this area \cite{liuOnlineAnomalousTrajectory2020, hanDeepTEAEffectiveEfficient2022,  zhang2023online, liCausalTADCausalImplicit2024}, key challenges persist—particularly the scarcity of labeled data, which limits the use of supervised methods. Labeling large-scale trajectory datasets is labor-intensive and often infeasible. While recent approaches have turned to unsupervised and self-supervised learning, reliably modeling normal patterns from unlabeled data remains difficult. In this work, we adopt contrastive learning to capture underlying movement regularities without requiring labeled anomalies.

Many existing methods classify an entire trajectory as anomalous or normal, offering limited interpretability and failing to localize anomalies. Fine-grained detection is more actionable—for example, pinpointing the exact segments where a taxi deviates from its expected route enables targeted interventions. Moreover, traditional methods relying on full trajectories struggle with mid-route detection. Our model addresses this by supporting sub-trajectory level and point-level anomaly detection, enabling flexible monitoring regardless of when the system is activated.
A further challenge arises from overlapping segments across different routes. In real-world networks, multiple routes may share common paths before diverging. A segment normal on one route may be anomalous on another, complicating detection. Our model explicitly captures route dependencies to reduce misclassification in such scenarios.
Noisy data is another key issue: normal and anomalous patterns often coexist within training data, making it hard to separate genuine anomalies from noise. Many existing models lack robustness and interpretability, limiting manual validation. Our approach clusters rare behaviors and includes a visualization module to support interpretable and user-refinable detection.

Lastly, many existing methods suffer from a strong dependence on predefined anomaly thresholds. A common approach involves computing an anomaly score for each trajectory and using a manually selected threshold to distinguish normal and anomalous patterns \cite{chenIBOATIsolationBasedOnline2013, zhangIBATDetectingAnomalous2011, wuFastTrajectoryOutlier2017, zhu2017effective, liuOnlineAnomalousTrajectory2020}. However, determining an optimal threshold is nontrivial, as it significantly impacts detection performance and generalization. Threshold-based approaches are also prone to overfitting to specific datasets. To address this issue, we introduce a threshold-free anomaly detection framework leveraging reinforcement learning techniques, which dynamically adjusts decision boundaries based on real-time trajectory variations.

\textbf{Contributions:}
In response to the challenges outlined earlier, this work introduces \modelname{}, an innovative \underline{C}ontrastive \underline{r}einforcement learning framework tailored for \underline{o}nline \underline{T}rajectory \underline{a}nomaly \underline{d}etection. \modelname{} represents a significant advancement by enabling both real-time (online) and retrospective (offline) anomaly detection at sub-trajectory and point levels, effectively addressing the limitations of traditional methods. Unlike existing approaches that depend on manually labeled datasets or rigid threshold-based scoring, \modelname{} leverages a fully self-supervised architecture, combining contrastive learning for extracting route patterns and reinforcement learning for classifying anomalies. This design empowers the framework to autonomously differentiate between normal and anomalous behaviors, eliminating the need for explicit labels or manual threshold adjustments, and significantly enhancing its adaptability to diverse scenarios. Furthermore, \modelname{} is designed to handle key challenges in trajectory data, including noise, route dependencies, and irregular sampling, ensuring robustness in complex real-world transportation environments. To enhance interpretability, the framework integrates advanced visualized clustering techniques, facilitating human-in-the-loop threshold calibration while preserving operational flexibility. Comprehensive experiments conducted on real-world trajectory datasets confirm the exceptional performance of \modelname{} in terms of anomaly detection accuracy, adaptability, and visualization quality, establishing it as a powerful solution for trajectory anomaly detection.

The key contributions of this paper are as follows:
\begin{itemize}
    \item We introduce a self-supervised trajectory anomaly detection framework that supports online and offline anomaly detection at both sub-trajectory and point levels. By leveraging contrastive and reinforcement learning, our method eliminates the reliance on labeled data, addressing the challenge of label scarcity.
    \item Unlike traditional threshold-based models, we formulate sub-trajectory anomaly detection as a reinforcement learning problem, where a learned policy classifies anomalies without the need for a fixed threshold. Additionally, we introduce a biased reward function to enhance robustness against out-of-distribution (OOD) cases, improving generalization across diverse datasets.
    \item We perform extensive experiments and case studies using real-world transportation datasets to evaluate \modelname{}'s effectiveness in identifying trajectory anomalies. Our results demonstrate state-of-the-art detection accuracy, superior adaptability to dynamic route variations, and high-quality anomaly visualizations.
\end{itemize}

\section{Related Work}

\subsection{Self-Supervised Learning}
Self-supervised learning (SSL) is a machine learning paradigm where models are trained on unlabeled data with some designed pretext tasks. It aims to produce descriptive and intelligible representations \cite{balestrieroCookbookSelfSupervisedLearning2023}.

A common approach in SSL involves reconstructing original information from the spatial or temporal contexts of distorted or masked inputs. For instance, BERT \cite{devlin-etal-2019-bert} employs masked language modeling, where the model predicts masked tokens within a sentence to enhance its generalization capabilities for downstream tasks. Similarly, in computer vision, some studies mask pixels or patches in images and train models to recover them, which significantly improves performance on downstream tasks \cite{he2022masked}. This paradigm can also be applied to spatial-temporal data. For example, to better capture the periodicity of traffic data in forecasting applications, STEP \cite{shao2022pre} masks patches from a relatively long time series and utilizes a masked auto-encoder to predict these missing patches. The forecasting performance is greatly enhanced with the use of the pre-trained model.

Contrastive learning is a prominent SSL technique that aims to maximize the similarity between representations of positive samples while minimizing the similarity between negative samples. Initially, this method involves one positive sample and one negative sample \cite{chechik2010large}. Over time, it has evolved to include multiple positive and negative samples in a batch during training \cite{ oord2018representation}. MoCo \cite{he2020momentum} and SimCLR \cite{simclr} are two influential paradigms in contrastive learning that have significantly advanced the development of modern approaches in trajectory representation learning (TRL) \cite{changContrastiveSimilarity2023}. Their methodologies provide a solid foundation for further research and applications. Additionally, contrastive learning has demonstrated effectiveness in out-of-distribution (OOD) detection \cite{guille2025cadet}, indicating its promising potential for applications in trajectory anomaly detection.

\subsection{Trajectory Anomaly Detection}
Trajectory anomaly detection identifies deviations from normal routes and specifies which parts of a trajectory are abnormal. Existing methods can be classified into heuristic and learning-based approaches.

\textbf{Heuristic Methods.} The heuristic methods usually measure deviations of trajectories by predefined metrics and set a threshold for labeling anomalies with domain expert knowledge. In TRAOD \cite{leeTrajectoryOutlierDetection2008}, the authors present a partition-and-detect framework in which trajectory segments are compared with reference ones. iBAT \cite{zhangIBATDetectingAnomalous2011} is an isolation-based method, and it proposes to compute a trajectory's anomaly score based on the calling times of an isolation procedure that aims to isolate the given trajectory from reference trajectories. The above methods cannot be applied to partially generated trajectories (in an online manner). CTSS \cite{zhangContinuousTrajectorySimilarity2022} proposes an online detection approach by comparing the reference route with a potential whole trajectory that results in the minimum discrete Frechet distance. However, it still cannot tell which part of a trajectory deviates from normal routes. Extending from the isolation strategy in iBAT, iBOAT \cite{chenIBOATIsolationBasedOnline2013} is an algorithm that aligns with the goal of detecting sub-trajectories online. However, comparing trajectories with a historical dataset remains required, reducing detection efficiency. Generally, the heuristic methods rely heavily on the designed anomaly metrics, yielding additional hyperparameters to be tuned. Because of comparing trajectories with historical data, they are less efficient and are normally coarse-grained.

\textbf{Learning-Based Methods.} Most recent work is learning-based due to the expressive power of machine learning models. ATD-RNN \cite{songAnomalousTrajectoryDetection2018} is an RNN-based model that is trained in a supervised manner. Owing to the unavailability of labels, training with labeled data is not prominent. DB-TOD \cite{wuFastTrajectoryOutlier2017} is a linear probabilistic model that can be applied to both complete and partially observed trajectories. However, the capability of modeling complex problems is limited by its linear nature. GM-VSAE \cite{liuOnlineAnomalousTrajectory2020} proposes to detect trajectory anomalies by measuring trajectory generation errors. It leverages a VAE-based structure, restricting the distribution of latent trajectory embeddings to a Gaussian mixture distribution. The algorithm can be applied to partially generated trajectories and detect trajectory-level anomalies online. Sharing the same basic architecture and detection strategy with GM-VSAE, DeepTEA \cite{hanDeepTEAEffectiveEfficient2022} stresses the impact of traffic conditions on routes and trajectories. Likewise, ATROM \cite{gaoOpenAnomalousTrajectory2023} uses this architecture to recognize unknown anomalous trajectories. Additionally, a recent study \cite{liCausalTADCausalImplicit2024} incorporates causal learning to enhance the model's generation capability on unseen trajectory data. The above-mentioned methods are built upon strong assumptions that can be inconsistent with real-world scenarios, resulting in limited performance. Meanwhile, although anomalies are detected online, they cannot tell which part of a trajectory is anomalous. RL4OASD \cite{zhang2023online} is the latest online sub-trajectory anomaly detection algorithm. However, the Markov assumption limits its ability to consider longer trajectory contexts. 

Unlike the above-mentioned algorithms, our method accurately identifies sub-trajectory-level deviations online and considers a longer traveling context. Additionally, it demonstrates robustness against noise in training data.

\section{Preliminaries and Problem Definition}
\begin{figure*}[ht]
    \centering
    \includegraphics[width=\textwidth]{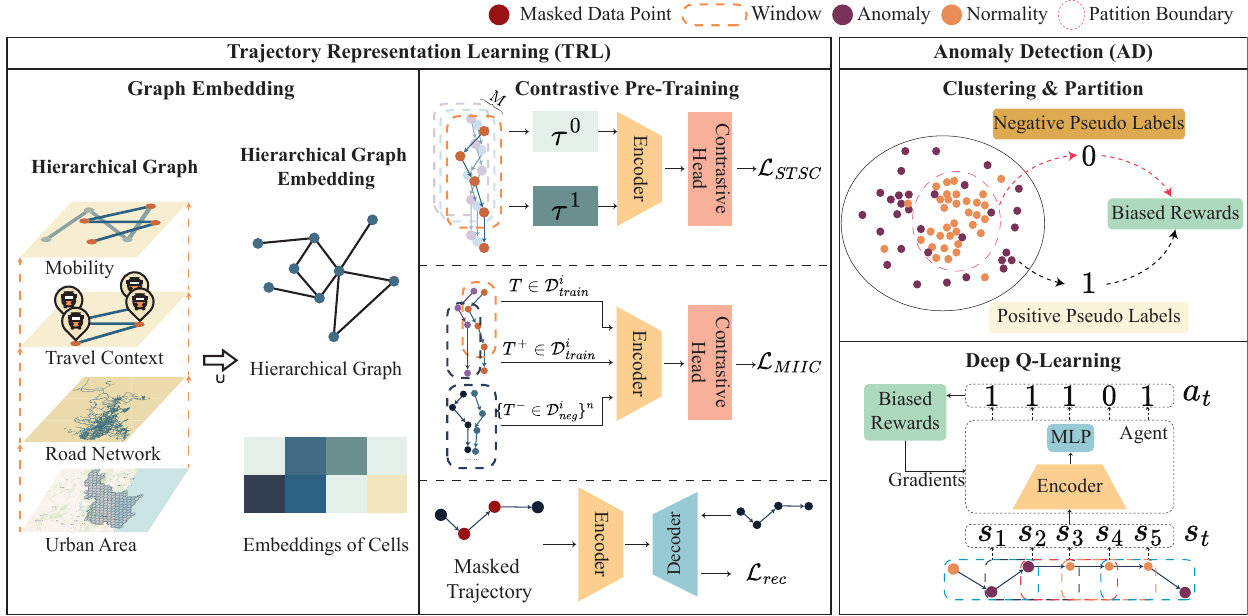}
    \caption{Overall framework of \modelname{}.}
    \label{fig:framework}
\end{figure*}
This section introduces key concepts forming the foundation for sub-trajectory anomaly detection and then formally defines the studied problem from multiple perspectives.
\subsection{Preliminaries}
\label{chap:preliminaries}
\begin{definition}[Trajectory] 
    A trajectory of length $n$, denoted by $T= \langle p_1, p_2, p_3, \dots, p_n \rangle$, is a chronologically ordered sequence of spatial points, representing the movement of an entity over time. Each point $p_k=(x_k,y_k,t_k)$ consists of the longitude $x_k$, latitude $y_k$, and the corresponding time $t_k$, indicating the entity's position at a specific timestamp.
\end{definition}

A \emph{sub-trajectory} is a segment of a trajectory that consists of a subsequence of consecutive spatial points derived from the original trajectory. Given a trajectory $T$ of length $n$, a sub-trajectory is denoted as $T_{i:j} = \langle p_i, p_{i+1}, p_{i+2}, \dots, p_j \rangle$ where $1\leq i \leq j \leq n$.

Modeling trajectories at the coordinate level is computationally expensive and impractical for extracting meaningful representations because of the high dimensionality and sparsity of the embedding space for precise GPS locations. To overcome this challenge, we represent each trajectory as a sequence of coarser-grained spatial units instead of using raw GPS coordinates, which is technically achieved through adopting the \textit{H3 Spatial Indexing System}\footnote{https://github.com/uber/h3}. This approach provides a hierarchical and efficient hexagonal tiling of the globe, significantly reducing the computational complexity while ensuring consistent and scalable spatial partitioning.   
A formal definition of H3-index cells is presented below.
\begin{definition}[H3-index Cell] An H3-index cell 
    is a discrete hexagonal region on a spherical surface, defined by the H3 spatial indexing system. Formally, an H3-index cell at resolution $\gamma$ is represented as: $c_\gamma = \{ l_1, l_2, l_3, l_4, l_5, l_6 \}$, where $c_\gamma$ represents a hexagonal region uniquely identified by the H3 index. Its boundary is collectively specified by six geographic coordinates $l_i = (x_i, y_i)$ where $1\leq i\leq 6$. Each cell at resolution $\gamma$ can be hierarchically subdivided into finer-resolution child cells at $\gamma+1$, preserving hierarchical relationships.
\end{definition}

Using the H3 spatial indexing system, we map-match each raw trajectory into a sequence of H3-index cells. Formally, given a trajectory $T = \langle p_1, p_2, \dots, p_n \rangle,$ 
the mapping process assigns each point to its corresponding H3-index cell at a specified resolution $\gamma$. This results in a sequence of cells: $T_c = \langle c_1, c_2, \dots, c_n \rangle$, where each cell is derived through $c_i = \text{H3}(p_i, \gamma)$. Here, $\text{H3}(p_i, \gamma)$ indicates an H3 cell at resolution $\gamma$ that contain the spatial point $p_i$. 
To eliminate redundancy and simplify representation, we remove consecutive duplicate cells in $T_c$, ensuring that each cell appears only once for a contiguous traversal.

\subsection{Problem Definition}
\label{chap:problem-definition}
We formulate the problems of offline and online sub-trajectory detection as follows.
\subsubsection{\textbf{Offline Sub-Trajectory Anomaly Detection}}
Let $\mathcal{T}_{od}$ be a set of historical trajectories associated with an origin location $o$ and destination location $d$. A sub-trajectory $T_{i:j}$ of a given trajectory $T$ associated with a pair of $od$ is considered \textit{anomalous} if it deviates significantly from the historical set of sub-trajectories derived from $\mathcal{T}_{od}$. Formally, a sub-trajectory $T_{i:j}$ is defined as anomalous with respect to $od$ if $\Pr(T_{i:j}|\mathcal{T}_{od}) \leq \delta$, where $\Pr(T_{i:j}|\mathcal{T}_{od})$ computes the probability of observing $T_{i:j}$ given the historical set $\mathcal{T}_{od}$, and $\delta$ is a predefined anomaly threshold. The threshold $\delta$ may be explicitly provided or inferred from the historical data. 


\subsubsection{\textbf{Online Sub-Trajectory Anomaly Detection}}

The online sub-trajectory anomaly detection extends the offline setting by employing a sliding window mechanism to evaluate sub-trajectories in real-time. To formalize the mechanism, consider a trajectory partially observed up to time $t$, denoted as $T_t = \langle p_1, p_2, \dots, p_t \rangle$. A sliding window of size $w$ is then applied to extract a sub-trajectory from $T_t$, that is, $T_{i:j} = \langle p_i, p_{i+1}, \dots, p_j \rangle$, where $i = t - w + 1$ and $j = t$. At the subsequent time step $t+1$, when a new point $p_{t+1}$ arrives, the sliding window shifts forward by one step and generates a new sub-trajectory: $T_{i+1:j+1} = \langle p_{i+1}, p_{i+2}, \dots, p_{j+1} \rangle$. This process continues iteratively as new points are observed.
Similarly, a sub-trajectory $T_{i:j}$ is considered \textit{anomalous} with respect to a pair of $od$ if it deviates significantly from the historical set of sub-trajectories derived from $\mathcal{T}_{od}$. 

\section{Methodology}
\subsection{Framework Overview}
\label{chap:overview}
This section introduces the methodology underlying our proposed model, which comprises two main components: trajectory representation learning and anomaly detection. The trajectory representation learning component transforms sub-trajectories into an embedding space where normal sub-trajectories are clustered closely together, and anomalous sub-trajectories are distinct, particularly from normal ones. The anomaly detection component identifies these anomalies using reinforcement learning enhanced with a novel biased reward mechanism to ensure robustness under OOD conditions. 

The trajectory representation learning module extracts structured embeddings of sub-trajectories. As illustrated in \figurename{~\ref{fig:framework}}, raw trajectories are first map-matched to an urban network, modelled as a hierarchical graph where nodes denote urban units and edges capture spatial and mobility relationships. Urban unit embeddings are then learned to encode this topology, allowing sequences of geospatial points to be represented as sequences of unit embeddings. To improve representation quality in the absence of labels, contrastive learning is developed using augmented sub-trajectories. Given the absence of labeled data, we apply several augmentation strategies to generate positive and negative samples for contrastive learning. Positive samples are sub-trajectories that share similar characteristics, while negative samples are generated by introducing variations that disrupt these similarities. The objective is that similar sub-trajectories cluster in the embedding space, while dissimilar ones are separated.

The anomaly detector combines offline clustering with reinforcement learning to identify abnormal sub-trajectories. Sub-trajectories are grouped in the embedding space, with dense clusters labelled as normal and sparse ones as potential anomalies. These pseudo-labels inform a biased reward function designed to promote detection accuracy and robustness to OOD patterns. A reinforcement learning agent learns from this feedback to refine the detection policy.

\subsection{Trajectory Representation Learning}
\subsubsection{Hierarchical Graph Embedding}
Topological information plays a crucial role in detecting trajectory anomalies. For example, transitions between two disconnected urban units can be flagged as anomalies, as such transitions are physically impossible. To encode this information into cells, we construct a \textbf{hierarchical graph} $\mathcal{G}_h(\mathcal{V}_h, \mathcal{E}_h)$ that represents relationships among urban units from various perspectives, leveraging multiple sources of information, including the road network, travel context, and historical mobility patterns.

\paragraph{Road Network Graph}
At the base of the topology lies the road network, which captures the physical connectivity of road segments in the real world. Nodes in the road network typically represent road intersections, while edges represent direct connections between these intersections. Let $\mathcal{G}_s(\mathcal{V}_s, \mathcal{E}_s)$ denote the road network of a given urban area. To integrate the road network into our framework, we perform a graph map-matching process to align it with the set of H3-index cells $U$, resulting in the map-matched graph $\mathcal{G}'_s(\mathcal{V}'_s, \mathcal{E}'_s)$, where $\mathcal{V}'_s \subseteq U$.

The mapping process $G$ is defined as:

\begin{equation}
\scalebox{0.8}{$
G:\begin{cases}
\forall v_i \in \mathcal{V}_s \Rightarrow \text{H3}(v_i) \in \mathcal{V}'_s, \\
\forall \langle v_i, v_j \rangle \in \mathcal{E}_s \land \text{H3}(v_i) \neq \text{H3}(v_j) \Rightarrow \langle \text{H3}(v_i), \text{H3}(v_j) \rangle \in \mathcal{E}'_s.
\end{cases}
$}
\end{equation}

\paragraph{Travel Context Graph}
In addition to the physical road network, travel context information is essential for certain applications. For instance, in public transportation systems, bus routes must adhere to predefined sequences of stops for a trip to be considered valid. 
Let $\mathcal{G}_c(\mathcal{V}_c, \mathcal{E}_c)$ represent the travel context graph, where nodes correspond to specific context points (e.g., bus stops), and edges encode sequential or functional relationships (e.g., the order of bus stops along a route). After map-matching, the resulting graph is denoted as $\mathcal{G}'_c(\mathcal{V}'_c, \mathcal{E}'_c)$, where $\mathcal{V}'_c \subseteq U$, and $\mathcal{G}'_c = G(\mathcal{G}_c)$.

\paragraph{Mobility Graph}
To capture common transition patterns between urban units, we construct a mobility graph $\mathcal{G}_m(\mathcal{V}_m, \mathcal{E}_m)$ based on historical trajectory data $\mathcal{D}_{train}$. For each trajectory $T_k \in \mathcal{D}_{train}$, where $1 \leq k \leq |\mathcal{D}_{train}|$, the mobility graph is constructed as follows:

\begin{equation}
\scalebox{0.8}{$
\mathcal{G}_m:\begin{cases}
\forall p_i \in T_k \Rightarrow p_i \in \mathcal{V}_m, \\
\langle p_j, p_{j+1} \rangle \in \mathcal{E}_m \iff p_j \in T_k \land p_{j+1} \in T_k \land T_{j:j+1} \subseteq T_k.
\end{cases}
$}
\end{equation}

Here, $p_i$ represents the $i$-th data point in trajectory $T_k$, and $T_{j:j+1}$ is a sub-trajectory of $T_k$. The map-matched mobility graph is denoted as $\mathcal{G}'_m(\mathcal{V}'_m, \mathcal{E}'_m) = G(\mathcal{G}_m)$.

\paragraph{Hierarchical Graph Construction}
To create a comprehensive hierarchical graph $\mathcal{G}_h$, we merge the road network, travel context graph, and mobility graph. The hierarchical graph is defined as:

\begin{equation}
\label{eq:graph-merge}
\mathcal{G}_h:\begin{cases}
\mathcal{V}_h = \mathcal{V}'_s \cup \mathcal{V}'_c \cup \mathcal{V}'_m, \\
\mathcal{E}_h = \mathcal{E}'_s \cup \mathcal{E}'_c \cup \mathcal{E}'_m.
\end{cases}
\end{equation}

To represent urban units (cells) in the embedding space, we use node2vec \cite{node2vec}, a widely used algorithm for learning node embeddings from graphs. Let $\boldsymbol{c}_i \in \mathbb{R}^{d_c}$ denote the embedding of the $i$-th cell in $U$. These embeddings encode the topological and contextual relationships captured in the hierarchical graph. 

\subsubsection{Contrastive Learning for Pre-Training}

Given the absence of labelled trajectory data, we employ contrastive learning as a self-supervised approach \cite{changContrastiveSimilarity2023, zhouGRLSTMTrajectorySimilarity2023, jiang2023self} to learn meaningful sub-trajectory representations. The primary objective of contrastive learning is to structure the embedding space such that sub-trajectories from similar movement patterns are mapped closer together, while dissimilar sub-trajectories are pushed apart. This ensures that normal sub-trajectories from frequently travelled routes form compact clusters, whereas anomalous sub-trajectories remain distinct and separable.

\paragraph{Sub-Trajectory-Similarity Contrast (STSC)}
Given a trajectory window \( T \) of fixed length \( L \), and the respective $od$, we randomly sample two augmentation functions, \( \mathcal{T}^0 \) and \( \mathcal{T}^1 \), from a predefined set of transformations \( \mathcal{T}_{pos} \). These functions are independently applied to \( T \), producing two augmented views, \( T_i = \mathcal{T}^0(T) \) and \( T_j = \mathcal{T}^1(T) \). Both views are passed through a shared encoder \( f \), yielding hidden representations:
$\boldsymbol{h}_i = f(T_i, od)$, $\boldsymbol{h}_j = f(T_j, od).$
To reduce potential information loss during contrastive learning, we apply a projection head \( g \) (typically a multi-layer perceptron) to map these hidden states into a lower-dimensional embedding space:
$\boldsymbol{z}_i = g(\boldsymbol{h}_i)$, $\boldsymbol{z}_j = g(\boldsymbol{h}_j).$
We optimize these embeddings using the normalized temperature-scaled cross-entropy loss (NT-Xent).
\begin{equation}
    \ell_{STSC}(i,j)=-\log{\frac{\exp(\frac{\operatorname{sim}(\boldsymbol{z}_i, \boldsymbol{z}_j)}{\tau})}{\sum_{k=1}^{2M} \mathbbm{1}_{\left[i \neq k\right]} \exp(\frac{\operatorname{sim}(\boldsymbol{z}_i, \boldsymbol{z}_k)}{\tau})}},
\end{equation}
where $\mathbbm{1}$ is an indicator function, $M$ is the batch size, and $\tau$ is a temperature hyperparameter. The function $\operatorname{sim}(\boldsymbol{z}_i, \boldsymbol{z}_j)$ represents cosine similarity:
\begin{equation}
    \operatorname{sim}(\boldsymbol{z}_i, \boldsymbol{z}_j) = \frac{\boldsymbol{z}_i \cdot \boldsymbol{z}_j}{\lVert \boldsymbol{z}_i \rVert \lVert \boldsymbol{z}_j \rVert}.
\end{equation}

The final contrastive learning objective, $\mathcal{L}_{STSC}$, is computed as the average NT-Xent loss across all pairs in a batch.
To ensure that sub-trajectories with similar mobility patterns are embedded closely together while dissimilar ones are pushed apart, we define a set of augmentation transformations $\mathcal{T}_{pos}$. These augmentations introduce controlled variations that mimic real-world trajectory distortions. The following transformations are used:

\begin{itemize}
\item\noindent
\textbf{Random Masking:}
Random masking removes a random subset of cells from a trajectory window \( T = \langle c_1, \dotsc, c_L \rangle \), simulating real-world data loss. The masked window is \( \tilde{T} = \langle c_{m_1}, \dotsc, c_{m_{\tilde{L}}} \rangle \), where the number of masked cells is drawn from \( \mathcal{U}(0, \rho_1) \). The hyperparameter \( \rho_1 \in [1, L] \) controls the masking rate.

\item\noindent
\textbf{Rear Truncation:}
Rear truncation removes the tail of a trajectory window \( T \) to align sub-trajectories with the same origin. The truncated version is \( \tilde{T} = \langle c_1, \dotsc, c_{\lceil m_{rt} \rceil} \rangle \), where \( m_{rt} \sim \mathcal{U}(0, \rho_2) \) and \( \rho_2 \in [1, L] \). This promotes robustness to temporary route deviations.


\item\noindent
\textbf{Head Truncation:}
Head truncation removes the front part of a window \( T \) to align sub-trajectories with the same destination:
$\tilde{T} = \langle c_{\lceil m_{ht} \rceil}, \dotsc, c_L \rangle, \quad m_{ht} \sim \mathcal{U}(0, \rho_3), \, \rho_3 \in [1, L].$
\end{itemize}


\paragraph{Intra-Itinerary Contrast (IIC)}

To structure the trajectory embedding space effectively, normal sub-trajectories should cluster densely while anomalies remain isolated. Intra-Itinerary Contrast (IIC) builds on the assumption that historical sub-trajectories within the same origin-destination (OD) pair in $\mathcal{D}_{train}$ are generally normal. However, two challenges arise: (1) sub-trajectories from different ODs may share segments (e.g., overlapping bus routes), making naive negative sampling unreliable; and (2) $\mathcal{D}_{train}$ may contain noisy or anomalous samples due to detours or GPS errors. To address this, we sample negatives of a trajectory window w.r.t a pair of $od$ from a curated domain $\mathcal{D}_{neg}^{od}$ and introduce a normality scoring function to better differentiate true anomalies from typical route variations. 

To measure the anomalousness of sub-trajectories, we introduce the frequently visited cells $S_{pos}^{od}$ and the infrequently visited cells $S_{neg}^{od}$. We first define the occurrence count of an element $x$ in a set $S$ as:
$C(x, S)=\sum_{i=1}^n\left[s_i=x\right], s_i \in S, |S| = n,$
where the Iverson bracket function $\left[s_i=x\right]$ is 1 if $s_i = x$, otherwise 0. Given a set of historical trajectories $\mathcal{D}_{train}^{od} \subseteq \mathcal{D}_{train}$ for the pair of $od$, the frequently visited set of cells is:
\begin{equation}
\label{eq:frequently_visited_cells}
    \scalebox{0.9}{$
    S_{pos}^{od} = \Set{c_i \given c_i \in U, \quad \frac{\sum_k C(c_i, T_k)}{|\mathcal{D}_{train}^{od}|} > \delta_{od}, \quad \forall T_k \in \mathcal{D}_{train}^{od}}$,
    }
\end{equation}
where $\delta_{od} \in (0,1]$ is a threshold controlling the acceptance of frequently visited cells. The complement set, containing infrequently visited cells, is $S_{neg}^{od} = U - S_{neg}^{od}.$

Accordingly, we define a normality scoring function:
\begin{equation}
\label{eq:normality score}
    \phi(T, od) = \frac{|T \cap S_{pos}^{od}|}{|T|},
\end{equation}
where $T$ is a map-matched sub-trajectory. The normality score $\phi(T, od)$ quantifies the proportion of frequently visited cells present in $T$ w.r.t the pair of $od$. Since $\mathcal{D}_{train}^{od}$ may contain noisy sub-trajectories, we must exclude these anomalies from the set of positive samples that define normal routes. The normality score offers a handle for differentiating noise. However, this metric alone is insufficient since it does not account for the order of cells in a trajectory, and thus can only serve as a rough noise filter, which is complementary to negative sampling. We integrate the normality score into a masked intra-itinerary contrastive loss $\mathcal{L}_{MIIC}$, which refines the set of normal sub-trajectories for each OD pair.
We define the masked intra-itinerary contrastive loss as follows:

\begin{gather}
\label{eq:MIIC main}
    \begin{split}
        \ell_{MIIC}^{od} = &\mathbb{E}_{T,T^+\sim\mathcal{D}_{train}^{od}}\\
        &\left(-\log\frac{\exp(x^+)}{\exp(x^+) + \mathbb{E}_{T^-\sim\mathcal{D}_{neg}^{od}}\bigl(\exp(x^-)\bigr)}\right)
    \end{split},\\
    x^+ = \lambda w_{T,T^+}^{od}\operatorname{sim}(\boldsymbol{z}, \boldsymbol{z}^+),\\
    x^- = \lambda\bigl(\operatorname{sim}(\boldsymbol{z},\boldsymbol{z}^-)+m\bigr),\\
    w_{T_1, T_2}^{od} = w\left(\min\left(\phi(T_1, od), \phi(T_2, od)\right)\right),\\
    \mathcal{L}_{MIIC} = \frac{1}{\left\lvert\mathcal{OD}\right\rvert} \sum_{od} \ell_{MIIC}^{od}.
\end{gather}

Here, $\boldsymbol{z} = f \circ g(T, od)$, $\boldsymbol{z}^+ = f \circ g(T^+, od)$, $\boldsymbol{z}^- = f \circ g(T^-, od)$, $\mathcal{OD}$ represents the set of all interested OD pairs. The parameter $m \in [0,2]$ controls embedding compactness, and $\lambda \geq 1$ defines the angular separability of embeddings in the hypersphere. The weighting function $w: \mathbb{R^{+}} \mapsto \mathbb{R}$ ensures that when a positive pair $(T, T^+)$ has a low normality score, it is treated as a negative pair.

To refine the selection of positive samples, we use the sign function:

\begin{equation}
w(s) = \begin{cases}
    1, & \text{if } s \geq \delta_1,\\
    0, & \text{if } \delta_2 < s < \delta_1,\\
    -1, & \text{otherwise}, 
\end{cases}
\end{equation}
where $\delta_1, \delta_2 \in [0,1]$ are tunable hyperparameters that adjust noise tolerance.
Since the normality score alone is insufficient for robust contrastive learning, we define the negative sample domain for $od$ as:

\begin{equation}
    \mathcal{D}_{neg}^{od} = (U^L - \mathcal{D}_{train}^{od}) \cup \mathcal{D}_{noise}^{od},
\end{equation}
where $\mathcal{D}_{noise}^{od} \subseteq \mathcal{D}_{train}^{od}$ contains noisy sub-trajectories. However, since $\mathcal{D}_{neg}^{od}$ is too large for efficient sampling, we approximate it using manually constructed negative samples, denoted as $\mathcal{D}_{neg'}^{od}$. Negative samples are generated using transformations $\mathcal{T} \in \mathcal{T}_{neg}$ applied to existing sub-trajectories. We introduce several transformations for constructing negative samples:
\begin{itemize}
\item \noindent
\textbf{Random Replacement:}
To simulate detours, we randomly replace a subset of cells in a sub-trajectory \( T \) with their \( k \)-hop neighbors from \( S_{neg}^{od} \), based on the road network or map-matching system. The number of replaced cells \( m \) is sampled uniformly from \( \{1, 2, \dots, L\} \), and the hop range \( k \in \mathbb{N} \) controls the perturbation extent.

\item \noindent
\textbf{Head and Rear Replacement:}
To simulate start or end detours, we replace the first or last \( m \in \{1, \dots, L\} \) cells of \( T \) with their \( k \)-hop neighbors in \( S_{neg}^{od} \), where \( k \in \mathbb{N} \) controls the perturbation range.

\item \noindent
\textbf{Negative Random Combination:}
To improve robustness to OOD cases, we construct negative samples \( T^- \in (S_{neg}^{od})^L \) by randomly sampling infrequently visited cells, as most anomalies involve such combinations.

\item \noindent
\textbf{Shuffling:}
To disrupt the inherent order of normal movement, we generate negative samples by randomly shuffling the cells in a trajectory \( T \), violating realistic travel patterns.

\item \noindent
\textbf{Repeating:}
Some sub-trajectories may appear normal in cell order but exhibit unrealistic patterns, such as loops. To simulate this, we generate a negative sample \( T^- = \langle c_{m_{L-k+1}}, \dots, c_{m_L}, c_{m_{L-1}}, \dots, c_{m_k} \rangle \) by reversing a segment of the original trajectory \( T = \langle c_{m_1}, \dots, c_{m_k}, \dots, c_{m_L} \rangle \).

\item \noindent
\textbf{Slices Permutation:}
Some anomalies arise from segment swaps within an otherwise valid trajectory. We simulate this by splitting \( T = \langle c_{m_1}, \dots, c_{m_k}, \dots, c_{m_L} \rangle \) at a random point \( c_{m_k} \) and swapping the two parts to form \( T^- = \langle c_{m_{k+1}}, \dots, c_{m_L}, c_{m_1}, \dots, c_{m_k} \rangle \).

\item \noindent
\textbf{Positive Random Combination:}
Even frequent cells can yield anomalies if arranged improperly. To capture this, we sample \( T^- \in (S_{pos}^{od})^L \), training the model to distinguish between structured and disordered normal movements.
\end{itemize}

\paragraph{Sub-trajectory Reconstruction}
To prevent model collapse and retain meaningful transition patterns, we incorporate a sub-trajectory reconstruction task during pre-training. Given a sub-trajectory \( T \), we generate a masked version \( \Tilde{T} \) by replacing a proportion of its cells with a special token \( \langle \texttt{mask} \rangle \), where the masking ratio is drawn from \( \mathcal{U}(0, \rho_4) \), with \( \rho_4 \in (0,1) \). Masking strategies include random, consecutive, or hybrid patterns.

The reconstruction model, denoted as $\Tilde{f}: \mathbb{R}^d \times U \mapsto \mathbb{R}^{|U|}$, learns to model the conditional probability of an urban unit given the preceding context:
\begin{gather}
    c_i \sim p(c \mid c_{<i}, od) = \operatorname{Mult}\left( \operatorname{softmax}\left( \boldsymbol{\hat{z}}_i\right)\right),\\
    \boldsymbol{\hat{z}}_i = \Tilde{f}(\boldsymbol{\hat{z}}_{i-1}, c_{i-1}), \quad \boldsymbol{\hat{z}}_0 = f(T, od),
\end{gather}
where $od$ represents the OD pair of sub-trajectory $T$, $c_i$ is the urban unit at time step $i$, and $c_0$ is initialized with a special token $\left\langle\texttt{start}\right\rangle$.

To optimize the reconstruction task, we define the reconstruction loss as:
\begin{equation}
    \mathcal{L}_{rec} = -\frac{1}{M} \sum_{i=1}^{M} \frac{1}{|T_i|} \sum_{j=1}^{|T_i|} \sum_{k=1}^{|U|} y_{i,j,k} \log p(c_k \mid c_{<j}, od),
\end{equation}
where $y_{i,j,k} = 1$ if urban unit $k$ is visited at time step $j$ in sub-trajectory $T_i$, and $y_{i,j,k} = 0$ otherwise. The batch size is denoted as $M$. Any applicable sequential modeling technique can be used for the decoder $\Tilde{f}$.

\paragraph{Pre-Training Objective}
The overall pre-training objective integrates the sub-trajectory reconstruction task with contrastive learning. The joint loss function is defined as:
\begin{equation}
    \mathcal{L}_{pre} = \alpha_1 \mathcal{L}_{rec} + \alpha_2 \mathcal{L}_{STSC} + \alpha_3 \mathcal{L}_{MIIC},
\end{equation}
where $\alpha_1, \alpha_2, \alpha_3$ are tunable weights controlling the contributions of the reconstruction loss ($\mathcal{L}_{rec}$), sub-trajectory similarity contrast loss ($\mathcal{L}_{STSC}$), and intra-itinerary contrast loss ($\mathcal{L}_{MIIC}$).

\subsubsection{Trajectory Representation Learning Module}

We assume that most normal sub-trajectories are permutations of frequently visited urban units. Given an OD pair \( od \), it is essential to distinguish between frequently visited cells \( S_{pos}^{od} \) and infrequently visited ones \( S_{neg}^{od} \). To enhance this distinction, we encourage the embeddings of cells in \( S_{od}^{pos} \) to be similar. To this end, we apply route-wise graph attention networks (GAT) \cite{veličković2018graph, brody2022how} on top of the base cell embeddings.

To mitigate mutual influence across OD pairs (e.g., shared segments in overlapping bus routes), we construct an individual subgraph \( \mathcal{G}_f^{od} \subseteq \mathcal{G}_h \) for each OD pair. This subgraph includes only frequently visited cells and their intra-connections:

\begin{equation}
\mathcal{G}_f^{od}(\mathcal{V}_f^{od}, \mathcal{E}_f^{od}) : 
\begin{cases}
\forall c_j \in S_{pos}^{od} \Rightarrow c_j \in \mathcal{V}_f^{od},\\
\langle c_j, c_k \rangle \in \mathcal{E}_f^{od} \iff 
\begin{aligned}
& c_j, c_k \in S_{pos}^{od},\\
& \langle c_j, c_k \rangle \in \mathcal{E}_h.
\end{aligned}
\end{cases}
\end{equation}


Let \( \boldsymbol{h}_j^{(l, od)} \in \mathbb{R}^{d_l} \) denote the embedding of cell \( c_j \) at layer \( l \), initialized with \( \boldsymbol{h}_j^{(0, od)} \in \mathbb{R}^{d_0} \). For simplicity, we drop the superscript \( l \) and \( od \) when not ambiguous. The attention coefficient between neighboring cells \( c_j \) and \( c_k \) is computed as:

\begin{align}
e(\boldsymbol{h}_j, \boldsymbol{h}_k) &= \boldsymbol{a}^\top \operatorname{LeakyReLU}(\boldsymbol{W}_1 [\boldsymbol{h}_j \| \boldsymbol{h}_k]), \\
\alpha_{jk} &= \frac{\exp(e(\boldsymbol{h}_j, \boldsymbol{h}_k))}{\sum_{k' \in \mathcal{N}_j^{od}} \exp(e(\boldsymbol{h}_j, \boldsymbol{h}_{k'}))},
\end{align}
where \( \boldsymbol{W}_1 \in \mathbb{R}^{d' \times 2d_{l-1}} \), \( \boldsymbol{a} \in \mathbb{R}^{d'} \), and \( \mathcal{N}_j^{od} \) is the neighbor set of cell \( c_j \) in \( \mathcal{G}_f^{od} \).
Using multi-head attention with \( H \) heads, the aggregated embedding at layer \( l \) becomes:

\begin{equation}
\boldsymbol{h}_j^{(l)} = \big\|_{m=1}^{H} \operatorname{activation} \left( \sum_{k \in \mathcal{N}_j^{od}} \alpha_{jk}^{(m)} \boldsymbol{W}_2^{(m)} \boldsymbol{h}_k^{(l-1)} \right),
\end{equation}
where \( \boldsymbol{W}_2^{(m)} \in \mathbb{R}^{\frac{d_l}{H} \times d_{l-1}} \) and \( \| \) denotes concatenation.
To retain structural information from the original hierarchical graph \( \mathcal{G}_h \), we combine the route-wise GAT output with the pre-trained cell embedding \( \boldsymbol{c}_j \). The final cell representation is:
$\boldsymbol{\hat{h}}_j^{od} = \operatorname{mlp}(\boldsymbol{h}_j^{(l)} \| \boldsymbol{c}_j)$, where \( \operatorname{mlp} \) denotes a feed-forward network. Each sub-trajectory \( T \) is then transformed into a sequence of cell embeddings: $T = \langle \boldsymbol{\hat{h}}_1^{od}, \boldsymbol{\hat{h}}_2^{od}, \dots, \boldsymbol{\hat{h}}_L^{od} \rangle$, which is fed into a sequential model. We use a recurrent neural network (RNN) initialized with the OD embedding \( \boldsymbol{r}_{od} \) to encode sub-trajectory dynamics: $\boldsymbol{\Tilde{H}} = \operatorname{RNN}(T, \operatorname{mlp}(\boldsymbol{r}_{od}))$,
where \( \boldsymbol{\Tilde{H}} = \langle \boldsymbol{\Tilde{h}}_1, \boldsymbol{\Tilde{h}}_2, \dots, \boldsymbol{\Tilde{h}}_L \rangle \in \mathbb{R}^{L \times d} \) are RNN outputs.
To capture the most informative points within the trajectory, we apply an attention mechanism over the RNN outputs, conditioned on the OD embedding:

\begin{equation}
\boldsymbol{h}^\top = \operatorname{mlp} \left( \operatorname{softmax} \left( \frac{(\boldsymbol{W}_q \boldsymbol{r}_{od})^\top (\boldsymbol{W}_k \boldsymbol{\Tilde{H}}^\top)}{\sqrt{d}} \right) \boldsymbol{\Tilde{H}} \right),
\end{equation}
where \( \boldsymbol{W}_q, \boldsymbol{W}_k \in \mathbb{R}^{d \times d} \) are learnable weights, and \( \boldsymbol{h} \in \mathbb{R}^d \) is the final representation of the sub-trajectory \( T \) with respect to the OD pair.

\subsection{Anomaly Detector}
In this section, we introduce the detector for our model, which include both offline and online detection modes. 
\subsubsection{Offline Detection}
Using the pre-trained trajectory representation learning model $f$, sub-trajectories are projected into an embedding space where normal patterns form dense clusters, while anomalies tend to lie in sparse regions. A straightforward approach to anomaly scoring in this space is to apply density estimation techniques \cite{germain2015made, papamakarios2017masked, dinh2017density, takahashiStudenttVariationalAutoencoder2018, pmlr-v80-huang18d, winkler2019learning, liuDensityEstimationUsing2021, marzouk2024distribution, patacchiola2024transformer}. Some prior works assume Gaussianity to model the distribution of embeddings \cite{lee2018simple, winkensContrastiveTrainingImproved2020, choMaskedContrastiveLearning2021}, but such assumptions may not hold in practice. Additionally, density-based scoring typically requires threshold tuning, which can be non-trivial and sensitive to the dataset.
To address these limitations, we adopt a clustering-based method for offline anomaly detection that avoids the need for explicit thresholding. Specifically, we apply the classical DBSCAN algorithm \cite{DBSCAN}, which is not only robust to outliers but also widely supported in modern libraries. Compared to threshold tuning in scoring-based methods, selecting DBSCAN’s hyperparameters is more intuitive, as they can be aligned with the degree of separation learned during pre-training.

Let $\mathcal{F}: \mathbb{N}^L \times \mathcal{OD} \rightarrow \{0, 1\}$ denote the binary labeling function that maps a map-matched sub-trajectory to its anomaly label, where $1$ indicates an anomaly and $0$ indicates normality. Given a set of historical sub-trajectories $\mathcal{D}_{train}^{od}$ for an associated OD pair $od$ and a test sub-trajectory $T$, we perform clustering on the combined set $\mathcal{D}_{train}^{od} \cup \{T\}$. Suppose DBSCAN assigns $T$ to a cluster $\mathcal{C}_T^{od}$ and partitions the data into $k$ clusters in total. We label $T$ as anomalous if it belongs to a relatively small cluster:

\begin{equation}
\mathcal{F}(T, od) = \begin{cases}
    1, & \text{if } |\mathcal{C}_T^{od}| < \frac{|\mathcal{D}_{train}^{od} \cup \{T\}|}{k},\\
    0, & \text{otherwise},
\end{cases}
\end{equation}
where $|\mathcal{C}_T^{od}|$ denotes the size of the cluster assigned to $T$. For computational efficiency, clustering can be performed on a sampled subset of $\mathcal{D}_{train}^{od}$ rather than the entire dataset.

\subsubsection{Online Detection}

Offline clustering-based detection is not suitable for real-time applications due to inefficiency and sensitivity to pseudo-label noise. To enable efficient and adaptive online detection, we introduce a Deep Q-Network (DQN)-based detection model \cite{mnih2015human}, which learns a detection policy from pseudo-labeled data.

\subsubsection{State and Action Design.}

For real-time detection, we apply a sliding window over the incoming trajectory. At each time step \(t\), the state is defined as $s_t = [T_{t:t+L-1}; od]$,
where \(T_{t:t+L-1} = \langle c_t, c_{t+1}, \dots, c_{t+L-1} \rangle\) is a sub-trajectory of length \(L\) and \(od\) denotes the origin–destination pair. The action space is \(A = \{0, 1\}\), where \(0\) indicates the sub-trajectory is normal and \(1\) indicates it is anomalous. At each time step, the model receives a new data point and forms a new state, enabling online labeling.

\subsubsection{Reward Function Design.}

To stabilize policy learning and improve robustness to out-of-distribution (OOD) inputs, we design a biased reward function that emphasizes correct anomaly detection. Let \(P\) and \(N\) denote the number of pseudo-labeled normal and anomalous sub-trajectories, respectively, in the training set \(\mathcal{D}_{train}^{od}\). We define the anomaly significance score as \(\theta = \frac{P}{N}\), reflecting the imbalance between normal and anomalous samples.

The reward function is structured to provide positive feedback when the agent’s prediction aligns with the pseudo-label and penalize incorrect predictions, with greater emphasis on detecting anomalies. The reward matrix is summarized in Table~\ref{tab:rewards}:

\begin{table}[ht]
    \centering
    \renewcommand{\arraystretch}{1.5}
    \caption{Biased Reward Function}
    \label{tab:rewards}
    \begin{tabular}{|l||c|c|}
    \hline
    \diagbox{Action}{Pseudo\\Label} & \(\mathcal{F}(s_t) = 0\) & \(\mathcal{F}(s_t) = 1\) \\
    \hline\hline
    \(a_t = 0\) & \(\frac{P+N}{P}\) & \(-\frac{P+N}{N} - \theta\) \\
    \hline
    \(a_t = 1\) & \(-\frac{P+N}{P}\) & \(\frac{P+N}{N} + \theta\) \\
    \hline
    \end{tabular}
\end{table}

This design ensures that the expected reward satisfies: $\mathbb{E}(R_t \mid a_t = 1) > \mathbb{E}(R_t \mid a_t = 0)$,
which encourages the agent to prefer anomaly labeling in uncertain situations. This aligns with practical expectations, as anomalies are both rare and critical to detect. While cluster size could be used to indicate anomalousness, we avoid this due to inconsistency: frequent detours may form large clusters, and some normal sub-trajectories may appear infrequently. Our reward function remains independent of cluster size and focuses solely on decision correctness and anomaly importance.

\subsubsection{Policy Learning.}

We fine-tune the trajectory representation learning model \(f\) along with a multi-layer perceptron (MLP) to approximate the Q-function: $Q(s_t, a_t) = \text{mlp}(f(s_t, od), a_t).$ The DQN loss is defined by the mean squared temporal difference (TD) error between the predicted and target Q-values:

\begin{multline}
\label{eq:dqn-loss}
\mathcal{L}_{DQN} = \frac{1}{|\mathcal{OD}|} \sum_{od} \mathbb{E}_{s_t \sim \mathcal{D}_{train}^{od}} \\
\Big[ R_{t+1} + \gamma \max_{a'} Q(s_{t+1}, a') - Q(s_t, a_t) \Big]^2,
\end{multline}
where \(R_{t+1}\) is the immediate reward and \(\gamma \in (0,1]\) is a discount factor.
Once trained, the detector selects the action that maximizes the Q-value for each new state: $a_t = \arg\max_{a'} Q(s_t, a').$

\subsubsection{Online Point-Level Detection}

Fine-grained anomaly detection is critical in practice, as it enables precise identification of anomalous segments within a trajectory. To this end, we propose a voting-based method for point-level detection, derived from sub-trajectory classification results.

Let the sub-trajectory window size be \(L\). When using a stride of 1, each point \(c_j \in T\) typically appears in \(L\) overlapping sub-trajectories, thus receiving up to \(L\) anomaly votes. However, the first and last \(L-1\) points of a trajectory appear in fewer windows. To ensure every point receives \(L\) votes, we pad the trajectory with variable-length windows at the beginning and end.
The first \(L-1\) states are constructed by gradually increasing the window size: $s_1 = [T_{1:1}; od], \dots, s_{L-1} = [T_{1:L-1}; od]$.
The last \(L-1\) states are created by decreasing the window size: $s_{|T|+1} = [T_{|T|-L+2:|T|}; od], \dots, s_{|T|+L-1} = [T_{|T|:|T|}; od]$.
The remaining \( |T| - L + 1 \) central states are formed by standard sliding windows of length \(L\), from \(s_L = [T_{1:L}; od]\) to \(s_{|T|} = [T_{|T|-L+1:|T|}; od]\). This padding strategy ensures that every data point in \(T\) receives exactly \(L\) anomaly votes, enabling uniform treatment across the trajectory.

Let \(\boldsymbol{\hat{a}} = \langle a_1, a_2, \dots, a_{|T|+L-1} \rangle\) be the binary action sequence produced by the anomaly detector. We define the vote aggregation function using a convolution with a kernel \(\boldsymbol{k}_{\mathbbm{1}} = \langle 1, 1, \dots, 1 \rangle\) of length \(L\):

\begin{equation}
\hat{\mathcal{F}}(\boldsymbol{\hat{a}})(j) = (\boldsymbol{\hat{a}} \star \boldsymbol{k}_{\mathbbm{1}})(j) = \sum_{n=0}^{L-1} \boldsymbol{\hat{a}}[j+n] \cdot \boldsymbol{k}_{\mathbbm{1}}[L - n],
\end{equation}
where \(\hat{\mathcal{F}}(\boldsymbol{\hat{a}})(j)\) denotes the anomaly score (vote count) assigned to the \(j\)-th point \(c_j\) in trajectory \(T\).
The final point-level anomaly label \(\hat{y}_j \in \{0, 1\}\) is determined by thresholding the vote count:

\begin{equation}
\hat{y}_j = 
\begin{cases}
1, & \text{if } \hat{\mathcal{F}}(\boldsymbol{\hat{a}})(j) \geq \delta_p, \\
0, & \text{otherwise},
\end{cases}
\end{equation}
where \(\delta_p \in \{1, 2, \dots, L\}\) is a tunable voting threshold. Setting \(\delta_p = L\) yields strict labeling, requiring unanimous anomaly votes.
As a post-processing step, we separately verify the first and last trajectory points \(c_1\) and \(c_{|T|}\). If either point does not belong to the frequently visited cell set \(S_{pos}^{od}\), it can be explicitly marked as anomalous, ensuring robust boundary detection.
This point-level voting strategy is applicable in both offline and online settings. In streaming scenarios, a data point \(c_j\) can be labeled as soon as the last of its \(L\) contributing sub-trajectories is observed.

\section{Experiments}
\begin{table*}[t]
    \centering
    \caption{Performance comparisons on two real-world datasets (higher is better; bold: best; underline: runner-up).}
    \resizebox{\textwidth}{!}{
    \begin{tabular}{l*{5}{|S[table-format=1.4, detect-weight]S[table-format=1.4, detect-weight]S[table-format=1.4, detect-weight]}}
        \hline
        \hline
        \multirow{3}{*}{Part I} & \multicolumn{15}{c}{\textit{Translink} (Window Level)}  \\
        \cline{2-16}
        & \multicolumn{3}{c|}{Head Detour} & \multicolumn{3}{c|}{Rear Detour} & \multicolumn{3}{c|}{Detour Midway} & \multicolumn{3}{c|}{Random Trajectory} & \multicolumn{3}{c}{Route Switching}\\
        \cline{2-16}
        & {P} & {R} & {F1} & {P} & {R} & {F1} & {P} & {R} & {F1} & {P} & {R} & {F1} & {P} & {R} & {F1}\\
        \cline{1-16}
        SAE  & 0.9943 & 0.9109 & 0.9508 & 0.9840 & 0.9072 & 0.9441 & 0.9931 & 0.9246 & 0.9577 & 1.0000 & 0.9994 & 0.9997 & 0.3048 & 0.0061 & 0.0119\\
        VSAE & 0.9978 & 0.9055 & 0.9494 & 0.9923 & 0.9035 & 0.9458 & 0.9961 & 0.9207 & 0.9569 & 1.0000 & 0.9993 & 0.9996 & 0.0758 & 0.0006 & 0.0012\\
        GM-VSAE & 0.8676 & 0.9825 & 0.9215 & 0.8474 & 0.9809 & 0.9092 & 0.9038 & 0.9834 & 0.9419 & 1.0000 & 1.0000 & 1.0000 & 0.5466 & 0.1934 & 0.2857\\
        SD-VSAE & 0.9979 & 0.9077 & 0.9507 & 0.9937 & 0.9029 & 0.9461 & 0.9973 & 0.9206 & 0.9574 & 1.0000 & 0.9993 & 0.9996 & 0.0874 & 0.0006 & 0.0011 \\
        DeepTEA & 0.9972 & 0.9259 & 0.9602 & 0.9926 & 0.9214 & \textbf{0.9557} & 0.9967 & 0.9357 & \underline{0.9653} & 1.0000 & 0.9996 & 0.9998 & 0.1461 & 0.0011 & 0.0021\\
        \cline{1-16}
        Clustering & 0.9755 & 0.9854 & \textbf{0.9804} & 0.9124 & 0.9920 & \underline{0.9505} & 0.9508 & 0.9887 & \textbf{0.9694} & 1.0000 & 1.0000 & \underline{1.0000} & 0.8767 & 0.9770 & \textbf{0.9242} \\
        \modelname{} & 0.9705 & 0.9848 & \underline{0.9776} & 0.8935 & 0.9907 & 0.9395 & 0.9399 & 0.9853 & 0.9620 & 1.0000 & 1.0000 & \textbf{1.0000} & 0.8757 & 0.9696 & \underline{0.9202} \\
        \hline
        \multirow{3}{*}{Part II} & \multicolumn{15}{c}{\textit{Translink} (Point Level)}  \\
        \cline{2-16}
        & \multicolumn{3}{c|}{Head Detour} & \multicolumn{3}{c|}{Rear Detour} & \multicolumn{3}{c|}{Detour Midway} & \multicolumn{3}{c|}{Random Trajectory} & \multicolumn{3}{c}{Route Switching}\\
        \cline{2-16}
        & {P} & {R} & {F1} & {P} & {R} & {F1} & {P} & {R} & {F1} & {P} & {R} & {F1} & {P} & {R} & {F1}\\
        \cline{1-16}
        SAE & 0.9982 & 0.5094 & 0.6746 & 0.9938 & 0.5073 & 0.6717 & 0.9973 & 0.5192 & 0.6829 & 1.0000 & 0.9669 & 0.9832 & 0.1328 & 0.0004 & 0.0009 \\
        VSAE & 0.9987 & 0.5119 & 0.6769 & 0.9948 & 0.5079 & 0.6725 & 0.9979 & 0.5193 & 0.6831 & 1.0000 & 0.9673 & 0.9834 & 0.0133 & 0.0000 & 0.0001 \\
        GM-VSAE & 0.8577 & 0.7570 & \underline{0.8042} & 0.8654 & 0.7560 & \underline{0.8049} & 0.8815 & 0.7597 & \underline{0.8161} & 1.0000 & 0.9848 & \underline{0.9924} & 0.4159 & 0.0772 & 0.1303 \\
        SD-VSAE & 0.9990 & 0.5150 & 0.6796 & 0.9962 & 0.5092 & 0.6740 & 0.9984 & 0.5205 & 0.6842 & 1.0000 & 0.9673 & 0.9834 & 0.0196 & 0.0000 & 0.0001 \\
        DeepTEA & 0.9983 & 0.5548 & 0.7132 & 0.9961 & 0.5700 & 0.7251 & 0.9985 & 0.5797 & 0.7335 & 1.0000 & 0.9733 & 0.9864 & 0.0424 & 0.0001 & 0.0002 \\
        \cline{1-16}
        Clustering & 0.5000 & 0.6511 & 0.5656 & 0.5006 & 0.6484 & 0.5650 & 0.5450 & 0.6768 & 0.6038 & 1.0000 & 0.6824 & 0.8113 & 0.5481 & 0.6855 & \underline{0.6092} \\
        \modelname{} & 0.9771 & 0.9791 & \textbf{0.9781} & 0.8768 & 0.9847 & \textbf{0.9276} & 0.9114 & 0.9763 & \textbf{0.9428} & 1.0000 & 1.0000 & \textbf{1.0000} & 0.8540 & 0.9015 & \textbf{0.8771} \\
        \hline
        \hline
        \multirow{3}{*}{Part I} & \multicolumn{15}{c}{\textit{Porto} (Window Level)}  \\
        \cline{2-16}
        & \multicolumn{3}{c|}{Head Detour} & \multicolumn{3}{c|}{Rear Detour} & \multicolumn{3}{c|}{Detour Midway} & \multicolumn{3}{c|}{Random Trajectory} & \multicolumn{3}{c}{Route Switching}\\
        \cline{2-16}
        & {P} & {R} & {F1} & {P} & {R} & {F1} & {P} & {R} & {F1} & {P} & {R} & {F1} & {P} & {R} & {F1}\\
        \cline{1-16}
        SAE & 0.8333 & 0.9572 & 0.8910 & 0.8627 & 0.9482 & 0.9034 & 0.8907 & 0.9504 & 0.9229 & 1.0000 & 0.9933 & 0.9967 & 0.6420 & 0.5289 & 0.5800  \\
        VSAE & 0.9375 & 0.9393 & 0.9384 & 0.9565 & 0.9259 & 0.9409 & 0.9657 & 0.9329 & 0.9490 & 1.0000 & 0.9933 & 0.9967 & 0.8325 & 0.3646 & 0.5071 \\
        GM-VSAE & 0.7613 & 0.9797 & 0.8568 & 0.7732 & 0.9764 & 0.8630 & 0.8271 & 0.9764 & 0.8955 & 1.0000 & 0.9986 & \underline{0.9993} & 0.6178 & 0.9190 & 0.7389 \\
        SD-VSAE & 0.9623 & 0.9302 & 0.9460 & 0.9806 & 0.9146 & 0.9465 & 0.9839 & 0.9205 & \underline{0.9507} & 1.0000 & 0.9911 & 0.9955 & 0.8181 & 0.1576 & 0.2643 \\
        DeepTEA & 0.9572 & 0.9413 & 0.9492 & 0.9700 & 0.9372 & \underline{0.9533} & 0.9784 & 0.9460 & \textbf{0.9619} & 1.0000 & 0.9954 & 0.9977 & 0.6517 & 0.0994 & 0.1725 \\
        \cline{1-16}
        Clustering & 0.9412 & 0.9715 & \textbf{0.9561} & 0.9398 & 0.9731 & \textbf{0.9562} & 0.9491 & 0.9389 & 0.9440 & 1.0000 & 0.9950 & 0.9975 & 0.9646 & 0.8980 & \underline{0.9301} \\
        \modelname{} & 0.9431 & 0.9570 & \underline{0.9500} & 0.9393 & 0.9661 & 0.9525 & 0.9463 & 0.9257 & 0.9359 & 1.0000 & 0.9994 & \textbf{0.9997} & 0.9474 & 0.9432 & \textbf{0.9453} \\
        \hline
        \multirow{3}{*}{Part II} & \multicolumn{15}{c}{\textit{Porto} (Point Level)}  \\
        \cline{2-16}
        & \multicolumn{3}{c|}{Head Detour} & \multicolumn{3}{c|}{Rear Detour} & \multicolumn{3}{c|}{Detour Midway} & \multicolumn{3}{c|}{Random Trajectory} & \multicolumn{3}{c}{Route Switching}\\
        \cline{2-16}
        & {P} & {R} & {F1} & {P} & {R} & {F1} & {P} & {R} & {F1} & {P} & {R} & {F1} & {P} & {R} & {F1}\\
        \cline{1-16}
        SAE & 0.8810 & 0.7085 & 0.7854 & 0.9272 & 0.6804 & 0.7848 & 0.9351 & 0.6778 & 0.7860 & 1.0000 & 0.8949 & 0.9445 & 0.5933 & 0.1630 & 0.2558  \\
        VSAE & 0.9576 & 0.7105 & \underline{0.8157} & 0.9752 & 0.6837 & 0.8039 & 0.9731 & 0.6981 & 0.8130 & 1.0000 & 0.9064 & 0.9509 & 0.8041 & 0.1314 & 0.2259 \\
        GM-VSAE & 0.7146 & 0.8936 & 0.7941 & 0.7439 & 0.8781 & 0.8054 & 0.7749 & 0.8817 & 0.8248 & 1.0000 & 0.9855 & \underline{0.9927} & 0.5771 & 0.8413 & \underline{0.6846} \\
        SD-VSAE & 0.9677 & 0.7026 & 0.8141 & 0.9877 & 0.6659 & 0.7955 & 0.9848 & 0.6754 & 0.8013 & 1.0000 & 0.9018 & 0.9483 & 0.7518 & 0.0523 & 0.0979 \\
        DeepTEA & 0.9673 & 0.6896 & 0.8052 & 0.9780 & 0.6946 & \underline{0.8123} & 0.9827 & 0.7120 & \underline{0.8257} & 1.0000 & 0.9117 & 0.9538 & 0.5825 & 0.0405 & 0.0757 \\
        \cline{1-16}
        Clustering & 0.5286 & 0.7755 & 0.6287 & 0.5131 & 0.7807 & 0.6192 & 0.4838 & 0.7663 & 0.5931 & 1.0000 & 0.8081 & 0.8939 & 0.5280 & 0.7640 & 0.6244 \\
        \modelname{} & 0.9124 & 0.9653 & \textbf{0.9381} & 0.9028 & 0.9740 & \textbf{0.9370} & 0.8724 & 0.9465 & \textbf{0.9079} & 1.0000 & 0.9989 & \textbf{0.9995} & 0.8990 & 0.9184 & \textbf{0.9086} \\
        \hline
        \hline
    \end{tabular}
    }
    \label{tab:effectiveness_comparison}
\end{table*}

\subsection{Experimental Setup}
\subsubsection{Datasets} 
We conduct experiments on two real-world datasets: \textit{Porto}\footnote{\url{https://www.kaggle.com/c/pkdd-15-predict-taxi-service-trajectory-i/data}} and \textit{Translink}\footnote{\url{https://translink.com.au/about-translink/open-data}}. The \textit{Porto} dataset contains GPS trajectories from 442 taxis in Porto, recorded every 15 seconds between 1 July 2013 and 30 June 2014. For evaluation, we select 10 frequently used OD pairs (based on H3 cells). 
The \textit{Translink} dataset is collected from GTFS real-time feeds, covering bus GPS trajectories across multiple Brisbane routes from 1 April to 27 April 2023. We select 10 bus routes with sufficient data. Unlike \textit{Porto}, \textit{Translink} trajectories are sampled irregularly and are noisier, but routes are more consistent and abundant. We treat each bus route as a distinct OD pair (and see the respective two directions as two different routes w.r.t. the same OD). In contrast, \textit{Porto} has fewer, more varied trajectories per OD pair.

\subsubsection{Labels}
\label{sec:synthetic-testing-data}

Unlike earlier studies that rely on manual labelling of anomalous trajectories \cite{zhangIBATDetectingAnomalous2011, chenIBOATIsolationBasedOnline2013, zhang2023online}, we adopt a synthetic generation strategy due to the high cost and scalability limitations of manual annotation. Since neither the \textit{Porto} nor \textit{Translink} datasets contain labelled anomalous sub-trajectories, we simulate anomalies in the test set for evaluation. Given two consecutive points $c_i$ and $c_{i+1}$ in a sub-trajectory, we ensure geographic consistency by enforcing $\operatorname{dist}(c_i, c_{i+1}) \leq k_{hop}$, where $\operatorname{dist}$ is the hop distance within the road network or indexing system. We generate five types of anomalies:

\begin{itemize}
    \item \textbf{Head detour.} A detour is introduced at the start of a trajectory. We randomly choose $1 < i \leq |T|$ and replace the prefix $T_{1:i-1}$ with an anomalous segment $T'$, ensuring continuity by enforcing $\operatorname{dist}(c'_{i-1}, c_i) \leq k_{hop}$.

    \item \textbf{Rear detour.} A detour is added at the end. Selecting $1 \leq i < |T|$ randomly, we replace $T_{i+1:|T|}$ with $T'$ such that $\operatorname{dist}(c_i, c'_1) \leq k_{hop}$.

    \item \textbf{Midway detour.} A segment in the middle is replaced. Two random indices $1 < i < j < |T|$ are chosen, and $T_{i:j}$ is replaced with $T'$ satisfying $\operatorname{dist}(c'_1, c_i) \leq k_{hop}$ and $\operatorname{dist}(c'_{|T'|}, c_{j+1}) \leq k_{hop}$.

    \item \textbf{Random trajectory.} The entire trajectory $T$ is replaced with a randomly generated $T'$ from $S_{neg}^{od}$ to simulate OOD scenarios.

    \item \textbf{Route switching.} Two trajectories $T_1$ and $T_2$ from different OD pairs are combined to simulate a route switch. A new trajectory is formed by concatenating $T_1^{(1:\lfloor\beta |T_1|\rfloor)}$ and $T_2^{(\lfloor(1-\beta)|T_2|\rfloor:|T_2|)}$, where the $T_2$ segment is labeled anomalous. The hyperparameter $\beta \in (0,1)$ controls the mixing ratio.
\end{itemize}

\subsubsection{Parameters and Metrics}
We split each dataset into $70\%$ for training, $10\%$ for validation, and $20\%$ for generating test sets. For each anomalous trajectory generation method, 500 trajectories are randomly sampled from each OD pair. In the \textit{Porto} dataset, we use H3 resolution 10 (cell area: $0.0150\ \text{km}^2$) and a window size of 5. For \textit{Translink}, due to longer and more fixed bus routes, H3 resolution is set to 9 (cell area: $0.1053\ \text{km}^2$), with a window size of 10 to balance granularity and computational cost. The maximum allowable distance between consecutive urban units in generated trajectories is set to $k_{hop} = 3$. For route switching, the proportion parameter $\beta$ is sampled from $\mathcal{U}(0.3, 0.7)$, following previous works \cite{liuOnlineAnomalousTrajectory2020, hanDeepTEAEffectiveEfficient2022}.

Models are evaluated at both \textbf{window} and \textbf{point} levels to provide a comprehensive understanding of performance. Noisy sub-trajectories are labeled as normal due to the lack of ground truth. All experiments are conducted on a machine with an Intel Core i9-13900K CPU, 64GB DDR5 RAM, and an NVIDIA RTX 4090 GPU, running Ubuntu 22.04 LTS.
We use Precision (P), Recall (R), and F1-score—standard metrics for binary classification—for detailed model evaluation.

\subsubsection{Baselines}
We compare \modelname{} with several publicly available online trajectory anomaly detection algorithms:

\begin{itemize}
    \item \textbf{SAE} \cite{liuOnlineAnomalousTrajectory2020}: A basic RNN-based seq2seq model trained using reconstruction loss. Anomaly scores are computed from reconstruction errors.
    
    \item \textbf{VSAE} \cite{liuOnlineAnomalousTrajectory2020}: A VAE-based seq2seq model where latent embeddings follow a Gaussian distribution.
    
    \item \textbf{GM-VSAE} \cite{liuOnlineAnomalousTrajectory2020}: Extends VSAE with Gaussian Mixture modeling in latent space to represent multiple route modes.
    
    \item \textbf{SD-VSAE} \cite{liuOnlineAnomalousTrajectory2020}: A lightweight version of GM-VSAE that infers route modes from source-destination pairs.
    
    \item \textbf{DeepTEA} \cite{hanDeepTEAEffectiveEfficient2022}: Builds on GM-VSAE by incorporating temporal traffic conditions for time-sensitive anomaly detection.
\end{itemize}

Note that these baselines are designed for online point-wise detection, where inputs begin at the source and scores are generated only for the latest point. To enable window-level evaluation, we apply a heuristic: if any point in a window is marked anomalous, the entire window is treated as anomalous. While this favors the baselines, it does not fully reflect the capabilities of \modelname{}, which is designed to detect arbitrary sub-trajectory anomalies. All baselines are tuned using a development set, and their grid resolutions are aligned with ours: $0.32455 \times 0.32455 \ \text{km}^2$ for \textit{Translink} and $0.12267 \times 0.12267 \ \text{km}^2$ for \textit{Porto}.

\subsection{Effectiveness Evaluation}

Evaluation results on the synthetic test sets are shown in Table~\ref{tab:effectiveness_comparison}, comparing the clustering-based offline algorithm and the RL-based online model \modelname{}. While the clustering approach achieves the highest window-level scores, \modelname{} consistently outperforms all baselines at the point level across both datasets. Notably, \modelname{} achieves superior point-level performance even on the small-scale dataset. This highlights the scalability and generalization ability of \modelname{} under realistic, data-intensive scenarios.

Interestingly, all baselines report strong window-level performance due to the lenient labelling rule (i.e., labelling a window as anomalous if any point is anomalous). However, their low recall at the point level reveals a limitation: generative models can accurately identify some anomalous points but fail to detect most, making them unsuitable for precise sub-trajectory anomaly detection. \modelname{}, in contrast, achieves strong performance at both window and point levels, demonstrating its effectiveness in sub-trajectory anomaly detection. It also shows superior robustness in OOD settings such as random trajectories and route switching, benefiting from biased reward design and graph attention mechanisms. Baselines, although performing reasonably on random trajectories, consistently fail in route-switching scenarios, indicating a reliance on local transition patterns and an inability to capture higher-level route semantics. Finally, baselines perform slightly better on the smaller \textit{Porto} dataset, suggesting their performance is sensitive to the spatial resolution (i.e., number of grids). Moreover, since real-world noise is unlabeled in the test data, it may affect the perceived performance of \modelname{}.

\subsection{Denoising Evaluation}
\begin{table}[t]
    \centering
    \begin{threeparttable}
    \caption{Denoising evaluation on Translink at point level (bold: best; underline: runner-up).}
    \begin{tabular}{lccc}
    \toprule
        Method & Precision & Recall & F1-Score\\
    \midrule
        SAE & 0.1996 & 0.1255 & 0.1541 \\
        VSAE & 0.2831 & 0.2584 & 0.2702 \\
        GMVSAE & 0.1849 & 0.1989 & 0.1916 \\
        SD-VSAE & 0.3405 & \underline{0.4612} & \underline{0.3918} \\
        DeepTEA & \underline{0.4935} & 0.2483 & 0.3304 \\
        \modelname{} & \textbf{0.8179} & \textbf{0.8300} & \textbf{0.8239} \\
    \midrule
    Improvement & $65.73\%\uparrow$ & $79.97\%\uparrow$ & $110.29\%\uparrow$ \\
    \bottomrule
    \end{tabular}
    \end{threeparttable}
    \label{tab:denoising_evaluation}
\end{table}

The detection algorithm should be able to identify noise in the dataset, as detours—such as buses leaving routes to refuel—do not reflect normal transit behaviour. To assess the denoising capability of \modelname{}, we leverage the availability of reference routes in the \textit{Translink} dataset. Human annotators unfamiliar with the algorithm were invited to label test set trajectories using official route maps from the Translink website\footnote{\url{https://jp.translink.com.au/plan-your-journey/timetables}} to avoid bias. Since bus routes are fixed, annotators only marked urban units along each reference route, and any trajectory points outside those units were treated as noise. The intersection of all annotators' labels was used as the final reference.

Table~\ref{tab:denoising_evaluation} shows that baseline methods fail to effectively filter noise from the data, whereas \modelname{} reliably extracts normal routes and outperforms all baselines by a substantial margin—achieving improvements of up to 110.29\%. Notably, the noise tolerance of our model can be flexibly adjusted via parameters such as the frequently visited cell set in Eq.~\eqref{eq:frequently_visited_cells}, the weighting function in Eq.~\eqref{eq:MIIC main}, and the clustering parameters (if used).

\subsection{Window Analysis}
\begin{figure*}[ht]
    \centering
    \subfloat[Performance evaluations with varying window sizes on synthetic testing data. ]{\includegraphics[width=0.32\linewidth]{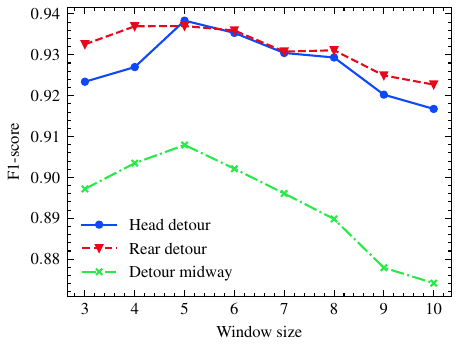}\label{fig:porto-varying-window-size}}
    \hfil
    \subfloat[Performance evaluations of different window sizes on synthetic window evaluation data with varying sub-trajectory lengths (tuned voting parameters for each window size).]{\includegraphics[width=0.32\linewidth]{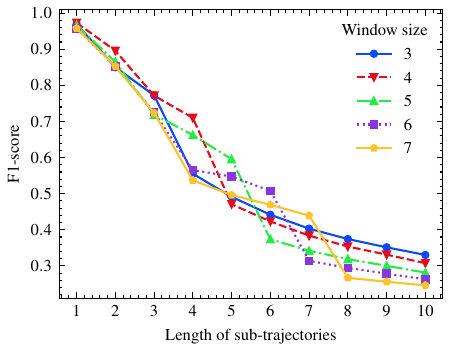}\label{fig:porto-window-eval-optimal-votes}}
    \hfil
    \subfloat[Performance evaluations of different window sizes on synthetic window evaluation data with varying sub-trajectory lengths (fixed voting parameters for all window sizes).]{\includegraphics[width=0.32\linewidth]{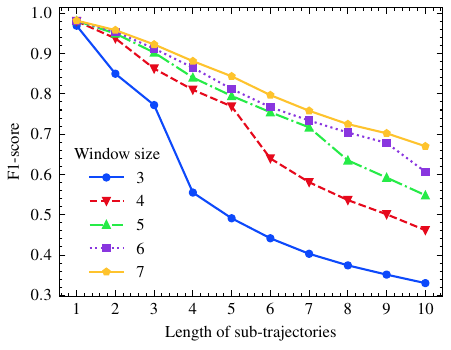}\label{fig:porto-window-eval-fixed-votes}}
    \caption{Window analysis on \textit{Porto} dataset at point level.}
\end{figure*}

To examine the effect of window size on point-level detection performance, we train models with varying window sizes on the smaller \textit{Porto} dataset. Results are shown in \figurename{~\ref{fig:porto-varying-window-size}}. Performance peaks at a window size of 5, suggesting that moderate-length windows provide an optimal balance between context and responsiveness. Notably, \modelname{} still performs reasonably well at smaller window sizes, indicating the feasibility of low-latency anomaly detection.

To further investigate the role of window size in capturing travel context, we create synthetic trajectories by stitching together multiple sub-trajectories of equal length from the same OD pair. Only the first sub-trajectory is labelled as normal, while the others are marked anomalous. This setup simulates abrupt deviations in otherwise normal routes. We then evaluate detection performance using fixed window sizes, as shown in \figurename{~\ref{fig:porto-window-eval-optimal-votes}} and \figurename{~\ref{fig:porto-window-eval-fixed-votes}}. As the sub-trajectory length increases, F1 scores decline due to the growing presence of normal-like patterns within anomalous segments. In \figurename{~\ref{fig:porto-window-eval-optimal-votes}}, a sharp drop occurs when the sub-trajectory length surpasses the window size, indicating that this size bounds the model’s contextual understanding. Similarly, in \figurename{~\ref{fig:porto-window-eval-fixed-votes}}, larger windows result in slower performance degradation, reinforcing their ability to capture broader context. These findings demonstrate that our model leverages contextual traveling information rather than memorizing local transitions alone, highlighting a key advantage over prior approaches.

\subsection{Ablation Study}
\begin{table*}[t]
    \centering
    \caption{Ablation Study on the Translink and Porto dataset (bold: best; underline: runner-up).}
    \resizebox{\textwidth}{!}{
    \begin{tabular}{l*{5}{|S[table-format=1.4, detect-weight]S[table-format=1.4, detect-weight]S[table-format=1.4, detect-weight]}}
        \toprule
        \hline
        \multirow{3}{*}{Part I} & \multicolumn{15}{c}{\textit{Translink} (Window Level)}  \\
        \cline{2-16}
        & \multicolumn{3}{c|}{Head Detour} & \multicolumn{3}{c|}{Rear Detour} & \multicolumn{3}{c|}{Detour Midway} & \multicolumn{3}{c|}{Random Trajectory} & \multicolumn{3}{c}{Route Switching}\\
        \cline{2-16}
        & {P} & {R} & {F1} & {P} & {R} & {F1} & {P} & {R} & {F1} & {P} & {R} & {F1} & {P} & {R} & {F1}\\
        \cline{1-16}
        w/o Graph Embedding & 0.9714 & 0.9647 & 0.9681 & 0.8880 & 0.9849 & 0.9339 & 0.9364 & 0.9768 & 0.9562 & 1.0000 & 1.0000 & 1.0000 & 0.8691 & 0.9776 & 0.9202 \\
        w/o route-wise GAT & 0.9660 & 0.9879 & 0.9768 & 0.8843 & 0.9924 & 0.9352 & 0.9347 & 0.9896 & 0.9614 & 1.0000 & 1.0000 & 1.0000 & 0.8614 & 0.9389 & 0.8985 \\
        w/o STSC & 0.9412 & 0.9872 & 0.9636 & 0.8658 & 0.9908 & 0.9241 & 0.9217 & 0.9881 & 0.9538 & 1.0000 & 1.0000 & 1.0000 & 0.8354 & 0.8636 & 0.8492 \\
        w/o MIIC & 0.8406 & 0.9396 & 0.8874 & 0.7980 & 0.9451 & 0.8653 & 0.8665 & 0.9444 & 0.9038 & 1.0000 & 0.9998 & 0.9999 & 0.6538 & 0.4507 & 0.5335 \\
        w/o Reconstruction & 0.9422 & 0.9720 & 0.9569 & 0.8692 & 0.9798 & 0.9212 & 0.9230 & 0.9763 & 0.9489 & 1.0000 & 0.9987 & 0.9994 & 0.8400 & 0.8547 & 0.8473\\
        Basic Rewards & 0.9713 & 0.9807 & 0.9760 & 0.8968 & 0.9852 & 0.9389 & 0.9415 & 0.9828 & 0.9617 & 1.0000 & 1.0000 & 1.0000 & 0.8758 & 0.9345 & 0.9042 \\
        Density Estimation & 0.9403 & 0.9965 & 0.9676 & 0.8625 & 0.9972 & 0.9250 & 0.9198 & 0.9960 & 0.9564 & 1.0000 & 1.0000 & 1.0000 & 0.8481 & 0.9917 & 0.9143 \\
        Density Estimation w/o PT & 0.6157 & 0.3018 & 0.4051 & 0.6029 & 0.2963 & 0.3973 & 0.7140 & 0.2937 & 0.4162 & 1.0000 & 0.2987 & 0.4600 & 0.6436 & 0.3030 & 0.4121 \\
        \cline{1-16}
        Clustering & 0.9755 & 0.9854 & \textbf{0.9804} & 0.9124 & 0.9920 & \textbf{0.9505} & 0.9508 & 0.9887 & \textbf{0.9694} & 1.0000 & 1.0000 & \underline{1.0000} & 0.8767 & 0.9770 & \textbf{0.9242} \\
        \modelname{} & 0.9705 & 0.9848 & \underline{0.9776} & 0.8935 & 0.9907 & \underline{0.9395} & 0.9399 & 0.9853 & \underline{0.9620} & 1.0000 & 1.0000 & \textbf{1.0000} & 0.8757 & 0.9696 & \underline{0.9202} \\
        \hline
        \hline
        \multirow{3}{*}{Part II} & \multicolumn{15}{c}{\textit{Translink} (Point Level)}  \\
        \cline{2-16}
        & \multicolumn{3}{c|}{Head Detour} & \multicolumn{3}{c|}{Rear Detour} & \multicolumn{3}{c|}{Detour Midway} & \multicolumn{3}{c|}{Random Trajectory} & \multicolumn{3}{c}{Route Switching}\\
        \cline{2-16}
        & {P} & {R} & {F1} & {P} & {R} & {F1} & {P} & {R} & {F1} & {P} & {R} & {F1} & {P} & {R} & {F1}\\
        \cline{1-16}
        w/o Graph Embedding & 0.9672 & 0.9603 & 0.9637 & 0.8664 & 0.9796 & 0.9196 & 0.8928 & 0.9688 & 0.9293 & 1.0000 & 1.0000 & \underline{1.0000} & 0.8514 & 0.9016 & 0.8758 \\
        w/o route-wise GAT & 0.9708 & 0.9831 & \underline{0.9769} & 0.8656 & 0.9875 & 0.9225 & 0.8979 & 0.9842 & 0.9391 & 1.0000 & 1.0000 & 1.0000 & 0.8325 & 0.8080 & 0.8201 \\
        w/o STSC & 0.9639 & 0.9829 & 0.9733 & 0.8547 & 0.9866 & 0.9160 & 0.8899 & 0.9811 & 0.9333 & 1.0000 & 1.0000 & 1.0000 & 0.7926 & 0.6660 & 0.7238 \\
        w/o MIIC & 0.9024 & 0.9225 & 0.9123 & 0.8194 & 0.9273 & 0.8700 & 0.8343 & 0.9169 & 0.8737 & 1.0000 & 0.9991 & 0.9996 & 0.4413 & 0.1650 & 0.2402 \\
        w/o Reconstruction & 0.9520 & 0.9448 & 0.9484 & 0.8653 & 0.9536 & 0.9073 & 0.8863 & 0.9479 & 0.9161 & 1.0000 & 0.9959 & 0.9979 & 0.8310 & 0.6977 & 0.7585 \\
        Basic Rewards & 0.9784 & 0.9718 & 0.9751 & 0.8830 & 0.9769 & \underline{0.9276} & 0.9158 & 0.9689 & \underline{0.9416} & 1.0000 & 1.0000 & 1.0000 & 0.8500 & 0.8176 & 0.8335 \\
        Density Estimation & 0.9437 & 0.9946 & 0.9685 & 0.8476 & 0.9951 & 0.9155 & 0.8642 & 0.9952 & 0.9251 & 1.0000 & 1.0000 & 1.0000 & 0.8397 & 0.9652 & \textbf{0.8981} \\
        Density Estimation w/o PT & 0.5395 & 0.3022 & 0.3874 & 0.5246 & 0.2970 & 0.3792 & 0.5759 & 0.2952 & 0.3904 & 1.0000 & 0.2988 & 0.4601 & 0.5448 & 0.3003 & 0.3872 \\
        \cline{1-16}
        Clustering & 0.5000 & 0.6511 & 0.5656 & 0.5006 & 0.6484 & 0.5650 & 0.5450 & 0.6768 & 0.6038 & 1.0000 & 0.6824 & 0.8113 & 0.5481 & 0.6855 & 0.6092 \\
        \modelname{} & 0.9771 & 0.9791 & \textbf{0.9781} & 0.8768 & 0.9847 & \textbf{0.9276} & 0.9114 & 0.9763 & \textbf{0.9428} & 1.0000 & 1.0000 & \textbf{1.0000} & 0.8540 & 0.9015 & \underline{0.8771} \\
        \hline
        \hline
        \multirow{3}{*}{Part I} & \multicolumn{15}{c}{\textit{Porto} (Window Level)}  \\
        \cline{2-16}
        & \multicolumn{3}{c|}{Head Detour} & \multicolumn{3}{c|}{Rear Detour} & \multicolumn{3}{c|}{Detour Midway} & \multicolumn{3}{c|}{Random Trajectory} & \multicolumn{3}{c}{Route Switching}\\
        \cline{2-16}
        & {P} & {R} & {F1} & {P} & {R} & {F1} & {P} & {R} & {F1} & {P} & {R} & {F1} & {P} & {R} & {F1}\\
        \cline{1-16}
        w/o Graph Embedding & 0.9378 & 0.9548 & 0.9462 & 0.9396 & 0.9546 & 0.9470 & 0.9458 & 0.9140 & 0.9297 & 1.0000 & 0.9995 & 0.9997 & 0.9432 & 0.9236 & 0.9333 \\
        w/o route-wise GAT & 0.9148 & 0.9671 & 0.9403 & 0.9338 & 0.9721 & 0.9525 & 0.9322 & 0.9315 & 0.9318 & 1.0000 & 0.9994 & 0.9997 & 0.9279 & 0.8382 & 0.8808 \\
        w/o STSC & 0.9469 & 0.9274 & 0.9370 & 0.9468 & 0.9280 & 0.9373 & 0.9533 & 0.8654 & 0.9072 & 1.0000 & 0.9986 & 0.9993 & 0.9525 & 0.9031 & 0.9271 \\
        w/o MIIC & 0.7933 & 0.8170 & 0.8049 & 0.8311 & 0.8070 & 0.8189 & 0.8262 & 0.7644 & 0.7941 & 1.0000 & 0.9777 & 0.9887 & 0.7610 & 0.5282 & 0.6236 \\
        w/o Reconstruction & 0.9196 & 0.9590 & 0.9389 & 0.9257 & 0.9606 & 0.9428 & 0.9328 & 0.9288 & 0.9308 & 1.0000 & 0.9984 & 0.9992 & 0.9140 & 0.7569 & 0.8281 \\
        Basic Rewards & 0.9604 & 0.9200 & 0.9398 & 0.9662 & 0.9124 & 0.9385 & 0.9674 & 0.8546 & 0.9075 & 1.0000 & 0.9914 & 0.9957 & 0.9706 & 0.7877 & 0.8696 \\
        Density Estimation & 0.8359 & 0.9945 & 0.9083 & 0.8446 & 0.9946 & 0.9135 & 0.8556 & 0.9846 & 0.9156 & 1.0000 & 1.0000 & \textbf{1.0000} & 0.8491 & 0.9985 & 0.9178  \\
        Density Estimation w/o PT & 0.5962 & 0.4638 & 0.5217 & 0.5843 & 0.4631 & 0.5167 & 0.6404 & 0.4603 & 0.5356 & 1.0000 & 0.5012 & 0.6678 & 0.6169 & 0.5052 & 0.5555 \\
        \cline{1-16}
        Clustering & 0.9412 & 0.9715 & \textbf{0.9561} & 0.9398 & 0.9731 & \textbf{0.9562} & 0.9491 & 0.9389 & \textbf{0.9440} & 1.0000 & 0.9950 & 0.9975 & 0.9646 & 0.8980 & \underline{0.9301} \\
        \modelname{} & 0.9431 & 0.9570 & \underline{0.9500} & 0.9393 & 0.9661 & \underline{0.9525} & 0.9463 & 0.9257 & \underline{0.9359} & 1.0000 & 0.9994 & \underline{0.9997} & 0.9474 & 0.9432 & \textbf{0.9453} \\
        \hline
        \hline
        \multirow{3}{*}{Part II} & \multicolumn{15}{c}{\textit{Porto} (Point Level)}  \\
        \cline{2-16}
        & \multicolumn{3}{c|}{Head Detour} & \multicolumn{3}{c|}{Rear Detour} & \multicolumn{3}{c|}{Detour Midway} & \multicolumn{3}{c|}{Random Trajectory} & \multicolumn{3}{c}{Route Switching}\\
        \cline{2-16}
        & {P} & {R} & {F1} & {P} & {R} & {F1} & {P} & {R} & {F1} & {P} & {R} & {F1} & {P} & {R} & {F1}\\
        \cline{1-16}
        w/o Graph Embedding & 0.9097 & 0.9649 & \underline{0.9365} & 0.9069 & 0.9626 & \underline{0.9339} & 0.8768 & 0.9371 & \underline{0.9059} & 1.0000 & 0.9989 & 0.9994 & 0.9018 & 0.9123 & 0.9070 \\
        w/o route-wise GAT & 0.8702 & 0.9776 & 0.9208 & 0.8907 & 0.9793 & 0.9329 & 0.8378 & 0.9487 & 0.8898 & 1.0000 & 0.9989 & 0.9995 & 0.8610 & 0.7794 & 0.8182 \\
        w/o STSC & 0.8828 & 0.9808 & 0.9292 & 0.8748 & 0.9790 & 0.9240 & 0.8263 & 0.9441 & 0.8813 & 1.0000 & 0.9993 & \underline{0.9997} & 0.8721 & 0.9475 & 0.9082 \\
        w/o MIIC & 0.7540 & 0.8150 & 0.7833 & 0.7990 & 0.8116 & 0.8053 & 0.7030 & 0.7552 & 0.7282 & 1.0000 & 0.9694 & 0.9844 & 0.7128 & 0.4769 & 0.5715 \\
        w/o Reconstruction & 0.9290 & 0.9254 & 0.9272 & 0.9347 & 0.9223 & 0.9285 & 0.9152 & 0.8842 & 0.8994 & 1.0000 & 0.9903 & 0.9951 & 0.9062 & 0.5520 & 0.6860 \\
        Basic Rewards & 0.9376 & 0.9148 & 0.9260 & 0.9361 & 0.9015 & 0.9185 & 0.9140 & 0.8585 & 0.8854 & 1.0000 & 0.9859 & 0.9929 & 0.9074 & 0.6655 & 0.7679 \\
        Density Estimation & 0.8574 & 0.9832 & 0.9160 & 0.8612 & 0.9831 & 0.9181 & 0.8080 & 0.9801 & 0.8858 & 1.0000 & 0.9995 & \textbf{0.9997} & 0.8739 & 0.9803 & \textbf{0.9240} \\
        Density Estimation w/o PT & 0.5337 & 0.4587 & 0.4933 & 0.5180 & 0.4568 & 0.4855 & 0.5000 & 0.4569 & 0.4775 & 1.0000 & 0.5009 & 0.6675 & 0.5408 & 0.4961 & 0.5175 \\
        \cline{1-16}
        Clustering & 0.5286 & 0.7755 & 0.6287 & 0.5131 & 0.7807 & 0.6192 & 0.4838 & 0.7663 & 0.5931 & 1.0000 & 0.8081 & 0.8939 & 0.5280 & 0.7640 & 0.6244 \\
        \modelname{} & 0.9124 & 0.9653 & \textbf{0.9381} & 0.9028 & 0.9740 & \textbf{0.9370} & 0.8724 & 0.9465 & \textbf{0.9079} & 1.0000 & 0.9989 & 0.9995 & 0.8990 & 0.9184 & \underline{0.9086} \\
        \hline
        \bottomrule
    \end{tabular}
    }
    \label{tab:ablation_study}
\end{table*}

To assess the contributions of individual components in \modelname{}, we conduct an ablation study comparing the full model with the following variants:
\begin{enumerate}
 \item \textbf{W/O Graph Embedding:} Replace pre-trained graph embeddings with randomly initialised vectors by a single-layer embedding.
\item \textbf{W/O route-wise GAT:} Remove the route-aware graph attention layer to assess its impact on robustness.
\item \textbf{W/O STSC:} Exclude the Spatio-Temporal Sub-Trajectory Clustering (STSC) strategy.
\item \textbf{W/O MIIC:} Omit the Mutual Intra-Itinerary Contrast (MIIC) strategy.
\item \textbf{W/O Reconstruction:} Remove the sub-trajectory reconstruction task from pre-training.
\item \textbf{Basic Rewards:} Use naive rewards ($R_{TP} = R_{TN} = 1$, $R_{FP} = R_{FN} = -1$), treating pseudo-labels as ground truths.
\item \textbf{Density Estimation:} Replace clustering with conditional neural spline flows \cite{durkan2019neural, winkler2019learning} to estimate $\Pr(T_{i:j}|\mathcal{D}_{train}^{od})$.
\item \textbf{Density Estimation w/o PT:} Apply the same density estimation directly to $\mathcal{D}_{train}$ without contrastive pre-training.
\item \textbf{Clustering:} Apply clustering-based detection alone (i.e., offline algorithm without DQN).
\end{enumerate}

The results in Table~\ref{tab:ablation_study} show that each component of \modelname{} contributes meaningfully to the final detection performance. The offline clustering-based method achieves the best window-level performance but lacks temporal continuity modeling, making it unsuitable for point-level detection. In contrast, \modelname{} performs best at the point level by leveraging reinforcement learning to model transitions between states, enabling fine-grained anomaly localization. It is also significantly more efficient than clustering in terms of time and memory. The density-estimation-based alternative performs competitively, especially in OOD scenarios such as random and route-switching anomalies, but its performance relies heavily on the proposed contrastive pre-training, without which its effectiveness drops substantially. This highlights the critical role of our pre-training strategy. Similarly, replacing our biased reward function with a naive -1/1 scheme degrades performance, particularly on smaller datasets, confirming that the proposed reward design better handles the rarity of anomalies and improves robustness. Lastly, the removal of the route-wise GAT results in performance drops, especially under noisy or rare-route conditions, demonstrating its importance in making embeddings more route-specific and robust. Overall, the study validates that each proposed component plays a vital role in enhancing anomaly detection performance, robustness, and interpretability.

\begin{figure}[t]
    \centering
    \subfloat[The reference route of Route 130 in Brisbane.]{\includegraphics[width=0.48\linewidth]{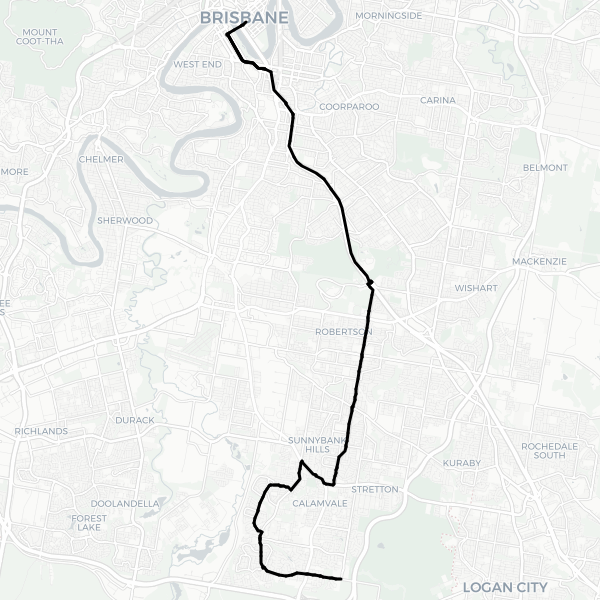}\label{fig:route-130-reference}}
    \hfil
    \subfloat[A sub-trajectory anomaly detection showcase of Route 130 detected by \modelname{}.]{\includegraphics[width=0.48\linewidth]{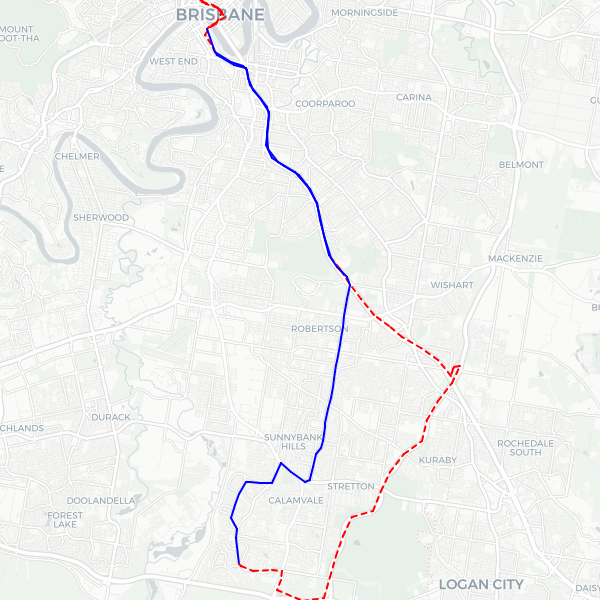}\label{fig:route-130-detection-case}}
    \\
    \subfloat[The reference route of Route 100 in Brisbane.]{\includegraphics[width=0.48\linewidth]{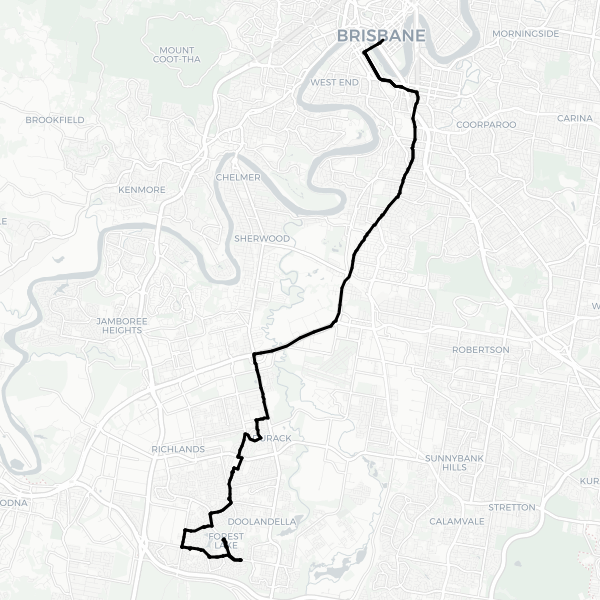}\label{fig:route-100-reference}}
    \hfil
    \subfloat[A sub-trajectory anomaly detection showcase of Route 100 detected by \modelname{}.]{\includegraphics[width=0.48\linewidth]{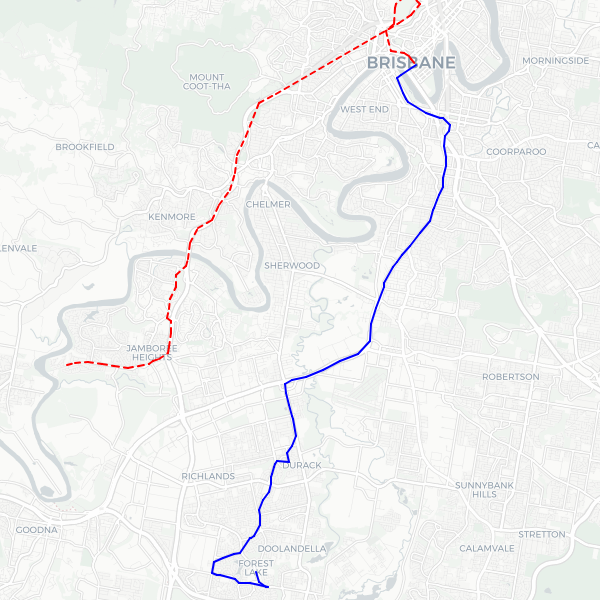}\label{fig:route-100-detection-case}}
    \caption{Sub-trajectory anomaly detection showcases on \textit{Translink}. Black lines are reference routes, blue lines denote normal sub-trajectories, and red dashed lines are anomalous sub-trajectories.}
    \label{fig:detection-case} 
\end{figure}

\begin{figure}[t]
    \centering
    \subfloat[Five hundred anomalous windows of Route 130 in \textit{Translink} labeled by clustering.]{\includegraphics[width=0.32\linewidth]{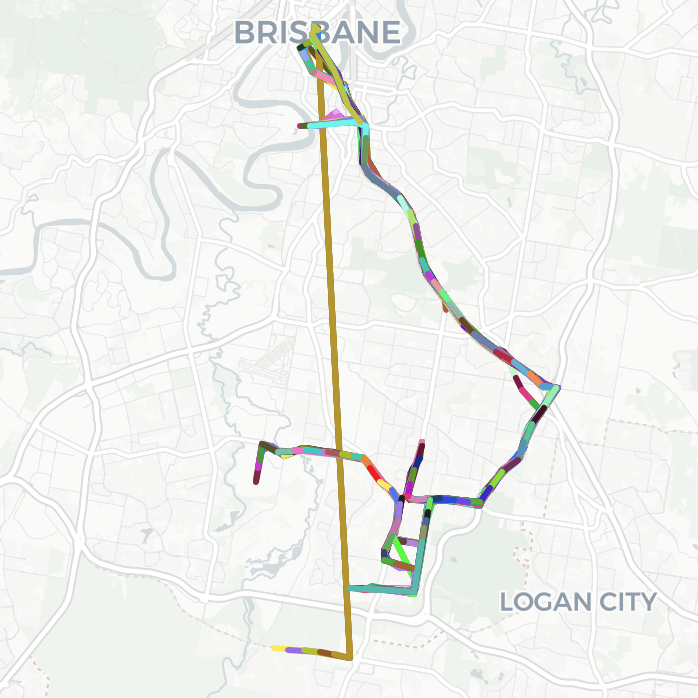}\label{fig:route-130-anomalous_cluster}}
    \hfil
    \subfloat[Five hundred normal windows of Route 130 in \textit{Translink} labeled by clustering.]{\includegraphics[width=0.32\linewidth]{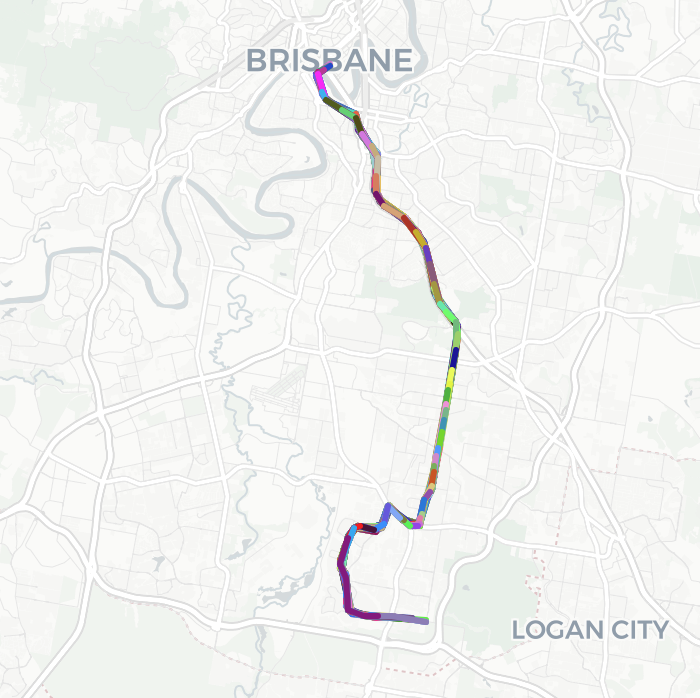}\label{fig:route-130-normal-cluster}}
    \hfil
    \subfloat[Five hundred normal windows of Route 130 in \textit{Translink} labeled by \modelname{}.]{\includegraphics[width=0.32\linewidth]{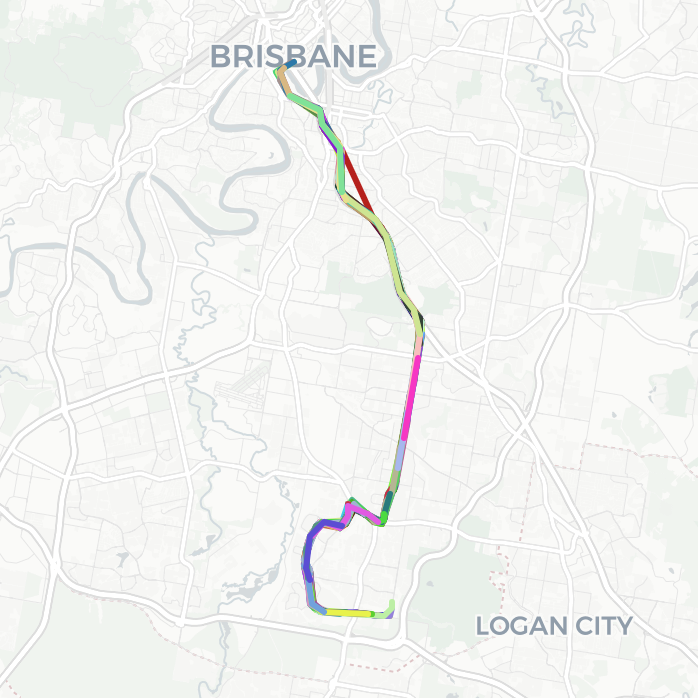}\label{fig:route-130-dqn-normal}}
    \\
    \subfloat[Five hundred anomalous windows of an OD in \textit{Porto} labeled by clustering.]{\includegraphics[width=0.32\linewidth]{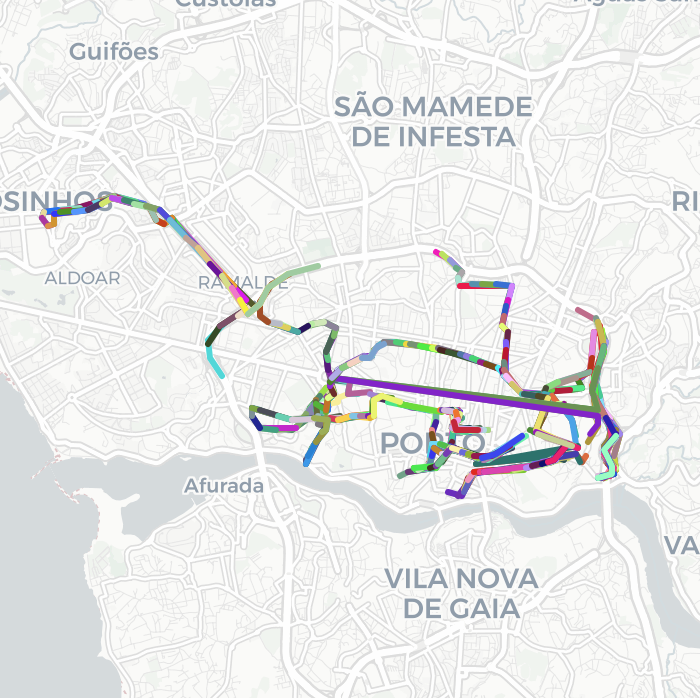}\label{fig:porto-anomalous_cluster}}
    \hfil
    \subfloat[Five hundred normal windows of an OD pair in \textit{Porto} labeled by clustering.]{\includegraphics[width=0.32\linewidth]{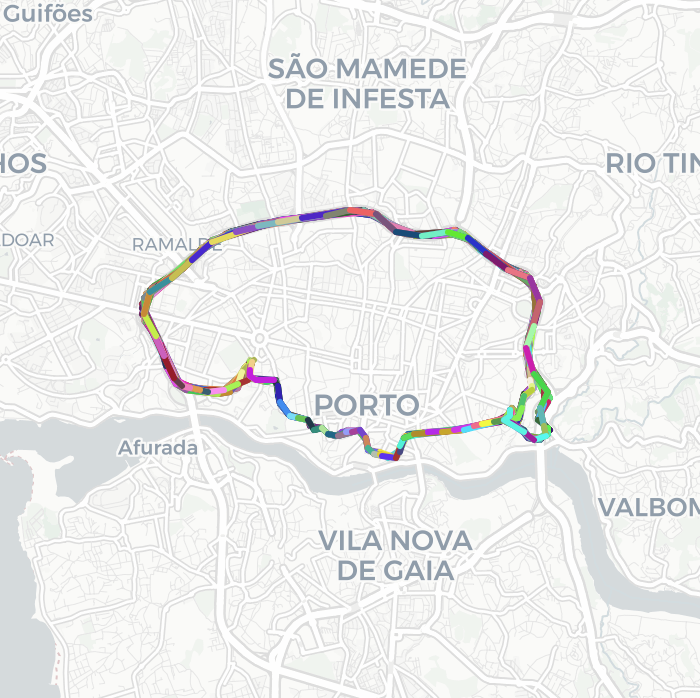}\label{fig:porto-normal-cluster}}
    \hfil
    \subfloat[Five hundred normal windows of an OD pair in \textit{Porto} labeled by \modelname{}.]{\includegraphics[width=0.32\linewidth]{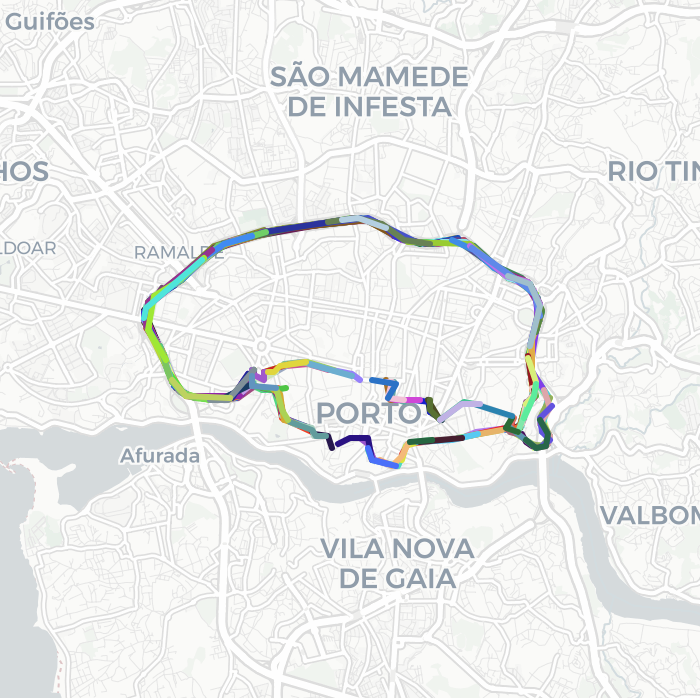}\label{fig:porto-dqn-normal}}
    \caption{Showcases of clustering results and extracted routes of different OD pairs on \textit{Translink} and \textit{Porto} by \modelname{}.}
    \label{fig:clustering-and-dqn-case} 
\end{figure}
\subsection{Case Study}
We provide illustrative examples using OpenStreetMap\footnote{\url{https://www.openstreetmap.org}} to visually demonstrate the effectiveness and interpretability of \modelname{}. In particular, \figurename{~\ref{fig:detection-case}} presents two detection scenarios from the \textit{Translink} dataset, where official reference routes are available for validation. \figurename{~\ref{fig:route-130-reference}} and \figurename{~\ref{fig:route-100-reference}} show the reference routes for Route 130 and Route 100, respectively, while \figurename{~\ref{fig:route-130-detection-case}} and \figurename{~\ref{fig:route-100-detection-case}} display the corresponding detection results. These figures highlight \modelname{}’s ability to identify anomalous sub-trajectories that deviate from expected paths accurately.

Further interpretability is achieved through our contrastive pre-training phase, which facilitates the clustering of normal sub-trajectories while dispersing anomalous ones into smaller or more isolated clusters. \figurename{~\ref{fig:clustering-and-dqn-case}} illustrates this behaviour: \figurename{~\ref{fig:route-130-anomalous_cluster}} and \figurename{~\ref{fig:porto-anomalous_cluster}} show examples of trajectory windows from smaller clusters, typically sparse and irregular segments not aligned with any known route, hence considered noisy. In contrast, \figurename{~\ref{fig:route-130-normal-cluster}} and \figurename{~\ref{fig:porto-normal-cluster}} depict windows from larger, denser clusters that effectively represent valid route segments. These examples confirm the pre-training strategy's capability to distinguish structured normal patterns from outliers. Finally, \figurename{~\ref{fig:route-130-dqn-normal}} and \figurename{~\ref{fig:porto-dqn-normal}} display windows identified as normal by the DQN agent in \modelname{}, reinforcing the model’s strength in learning and generalizing valid urban mobility behaviors.

\section{Conclusion}
This paper introduces a novel framework for sub-trajectory anomaly detection that integrates contrastive learning with deep reinforcement learning methodologies. We propose a robust pre-training phase designed to encourage the clustering of normal sub-trajectories in the embedding space while simultaneously pushing the representations of noisy sub-trajectories farther apart or forming smaller clusters. Building upon this pre-training, we develop a clustering-based offline detection algorithm that is both threshold-free and intuitive, thereby enhancing visualization capabilities and interpretability.

To facilitate efficient and nuanced anomaly detection, we present an online sub-trajectory anomaly detector, denoted as \modelname{}, which is trained using a biased reward function designed to improve robustness against OOD cases. Our extensive experimental evaluations substantiate the effectiveness of our proposed model. Notably, the clustering-based offline algorithm demonstrates superior performance in window-level detection, while \modelname{} excels in point-level detection.
Moreover, when juxtaposed with previous methodologies, our framework exhibits significant advantages in denoising capabilities. The visualization outcomes reveal that our approach is proficient in extracting normal routes without necessitating real-world labels, underscoring its practical applicability in diverse contexts.

\bibliographystyle{spbasic}
\bibliography{ref}

\end{document}